\documentclass[]{agujournal2019}
\usepackage{url} 

\usepackage{soul}

\usepackage{multicol}
\usepackage{stackengine}
\usepackage{physics}
\usepackage{tabularx}
\usepackage{graphicx}
\usepackage{amsmath}
\usepackage{cleveref}
\usepackage{booktabs}
\usepackage{bm}
\usepackage{xcolor}
\usepackage{amssymb}
\usepackage{comment}
\usepackage{lineno}

\def\vtx{\vb{\tilde{x}}}
\def\vx{\vb{x}}

\newcommand{\bs}[1]{\boldsymbol{#1}}

\makeatletter
\def\ps@headings{%
  \def\@oddfoot{\centerline{\small --\the\c@page--}}%
  \let\@evenfoot\@oddfoot
  \def\@oddhead{}%
  \let\@evenhead\@oddhead
}
\ps@headings
\makeatother

\draftfalse
\journalname{}

\begin{document}

%
%


\title{Reduced Order Modeling for Tsunami Forecasting with Bayesian Hierarchical Pooling}

%
%




\authors{Shane X. Coffing \affil{1}, John Tipton \affil{2}, Arvind T. Mohan \affil{1}, Darren Engwirda \affil{3}\affil{4}}


\affiliation{1}{Computing and Artificial Intelligence Division (CAI-2), Computational Physics and Methods, Los Alamos National Laboratory, Los Alamos, NM, USA}
\affiliation{2}{Computing and Artificial Intelligence Division (CAI-4), Statistics, Los Alamos National Laboratory, Los Alamos, NM, USA}
\affiliation{3}{Theoretical Division (T-3), Fluid Dynamics and Solid Mechanics, Los Alamos, Los Alamos, NM, USA}
\affiliation{4}{Environment, Commonwealth Scientific \& Industrial Research Organisation, Hobart, Australia}




\correspondingauthor{Shane X. Coffing}{sxc@lanl.gov}



\begin{keypoints}
\item Standard Galerkin-projection-based reduced-order models (GP-ROMs) may lose reliability when applied to scenarios with unseen dynamics.
\item We introduce a corrected GP-ROM with Bayesian hierarchical pooling for probabilistic calibration of tsunami forecasting scenarios.
\item We demonstrate that a random-coefficient ROM (randPROM) enables uncertainty-aware tsunami forecasting with fewer simulations.
\end{keypoints}

%
%

%
%


\begin{abstract}
Reduced-order models (ROMs) can represent spatiotemporal processes in significantly fewer dimensions and can often be solved many orders of magnitude faster than their governing partial differential equations (PDEs). For example, proper orthogonal decomposition yields a ROM in which the state is represented as a low-dimensional linear combination of fixed spatial modes and time-dependent coefficients, but this representation remains constrained by the process used to construct the basis. In this work, we explore a new type of ROM that is not restricted to a single fixed coefficient trajectory. Specifically, we consider a corrected Galerkin-projection ROM, formulated as an initial value problem that encodes the physics of the governing PDEs and is calibrated through operator corrections to more accurately reproduce the coefficient dynamics. By combining this corrected reduced model with a Bayesian hierarchical pooling framework over the initial reduced coefficients, we obtain new, statistically interpretable and physically grounded coefficient trajectories that generalize across related scenarios. When recombined with the spatial modes, these trajectories define a complete probabilistic physics surrogate, called a randPROM, for generating simulations that are distributionally consistent with a neighborhood of initial conditions near those used to construct the ROM. We apply the randPROM framework to tsunami modeling, a setting involving unpredictable, catastrophic, and strongly nonlinear dynamics, using both a synthetic case study near Fiji and the real-world 2011 Tohoku tsunami. We demonstrate that randPROMs can substantially reduce the number of full simulations required while providing statistically calibrated and physically defensible predictions of tsunami wave arrival times and heights.
\end{abstract}

\section*{Plain Language Summary}
Scientists often must simulate complex sets of equations to explain physical phenomenon, such as tsunami wave propagation. But due to modeling difficulties and long simulation times, these simulations are difficult to deploy for explaining the risks of tsunamis. We often have to rely on reduced order models (ROM), that are faster and simpler, but often fail to explain more general cases or variations in a model. We present a new ROM-based method, called a random-coefficient projection-based ROM (randPROM), that combines the strengths of fast solvers and Bayesian statistics, to create a rapid and flexible surrogate that can explain tsunami risks in near real-time. We demonstrate this model by predicting cases in Fiji as well as the real-world Tohoku 2011 disaster.

\section{Introduction}
High-fidelity models of geophysical flows require computational resources, detailed knowledge of initial conditions, and domain expertise that may exceed the timescales and practical forecasting needs of an emergency response. This is particularly true for unpredictable events such as earthquake-induced tsunamis, which require finely resolved numerical solutions of systems of partial differential equations (PDEs) to capture the highly nonlinear, nonstationary, and localized dynamics of a specific event. The shallow-water equations (SWEs) provide an applicable PDE system that reduces the computational complexity of a full three-dimensional Navier--Stokes solver and can be used to accurately model tsunami-wave propagation, in addition to many other oceanic and atmospheric flows. Tsunami waves are well approximated by the shallow-water equations because their wavelengths (on the order of hundreds of kilometers) greatly exceed the ocean depths through which they propagate (on the order of a few kilometers) \cite{titov1998numerical}. Despite their reduced computational complexity, the SWEs alone may still not satisfy the time constraints required to generate a full distribution of plausible scenarios. According to the National Oceanic and Atmospheric Administration, ``The main objective of a forecast model is to provide an estimate of wave arrival time, wave height and inundation area immediately after a tsunami event,'' with the key requirement being immediacy of the information if it is to serve as a useful decision-support tool during a crisis \cite{noaa}. Tools such as Tsunami Travel Time software \cite{wessel2009analysis} can provide rapid estimates of wave arrival times, but may provide only one component of the information required in practice.

Reduced-order models (ROMs) may also help meet the need for reliable, fast, and actionable information. These computationally efficient, low-dimensional representations of dynamical systems can provide approximations to the key characteristics of the problem of interest. A common mathematically based ROM is obtained through proper orthogonal decomposition (POD), also known as principal component analysis, which assumes that the dynamics of a system can be described by orthogonal spatial patterns, called modes, together with time-dependent coefficients. This technique has been used extensively in reduced-order modeling of hydrodynamic flows since its introduction \cite{lumley1979computational,berkooz1993proper}. A POD-based ROM is constructed by retaining a much smaller set of modes that still provides an accurate approximation to the full solution. A physically based ROM may then be obtained by projecting those retained POD modes back onto the governing PDE, thereby exploiting orthogonality and the algebraic structure of the equations to produce an ODE governing the temporal coefficients. The resulting model is referred to as a Galerkin-projection ROM (GP-ROM).

Despite their theoretical appeal, GP-ROMs may produce reduced dynamics that deviate qualitatively from those of the full system, particularly at low truncation ranks \cite{rowley2004model,callaham2022role}. Such behavior can include spurious equilibria or artificial limit cycles, even when the Galerkin system retains energy-stability properties inherited from the underlying PDEs. In practice, reduced-model error may also appear as phase drift, amplitude error, or long-time trajectory mismatch relative to the POD coefficients. To improve robustness, the GP-ROM operators may require calibration, preconditioning, or additional regularization, such as artificial viscosity terms that damp undesirable behavior \cite{bergmann2009enablers}. It has also been shown that reduced operators can be optimized to produce more stable and accurate ODE solutions \cite{mohan2021learning,chakrabarti2023full}. In this work, we extend that operator-correction perspective by calibrating both the linear and quadratic reduced operators, and by demonstrating successful training on problems requiring larger retained mode sets.

A remaining limitation of corrected GP-ROMs is that they are still designed to reproduce the coefficient dynamics associated with a given reduced dynamical system. While the underlying POD basis may be constructed from a broader set of simulations, the associated GP-ROM is still defined by a single set of reduced operators that may not accurately reproduce the dynamics of distinct trajectories within that broader subspace. In this work, we use a single simulation to construct the GP-ROM and its corrected reduced counterpart, while assuming that the resulting modes can represent a broader class of related unseen scenarios. Under this assumption, new coefficient trajectories can be combined with the modes to approximate those new scenarios.

Once a corrected reduced model is constructed, the next task is to supply it with physically meaningful initial reduced states. To do so, we perform statistical calibration through Bayesian hierarchical pooling \cite{gelman2006bayesian}. In the tsunami setting, we hypothesize that the spatial features of one tsunami model (a reference model) may also be used to describe another event (a test model) that is nearby in space and/or of similar earthquake magnitude, with acceptable accuracy. We use this framework to learn distributions over initial reduced coefficients that are consistent with observational data, and demonstrate the utility of random-coefficient projection-based reduced-order models (randPROMs) by calibrating these initial conditions against both simulated and real-world sensor data. In this way, the randPROM provides an interpretable probabilistic reduced model for each case. In this work, we show that randPROMs can reproduce key features of tsunami trajectories while also providing uncertainty quantification for related, previously unseen scenarios.

The remainder of the paper is organized as follows. Section 2 reviews the construction of a reduced-order model via POD, the corresponding GP-ROM from the SWE system, the operator-correction procedure used to improve the reduced dynamics, and the formulation of the randPROM framework. Section 3 applies that framework to a set of synthetic tsunami cases near Fiji. Section 4 applies the randPROM methodology to data from the 2011 Tohoku tsunami. Section 5 concludes the paper with a discussion of implications, limitations, and directions for future work.

\begin{figure}[t]
\includegraphics[width=\textwidth]{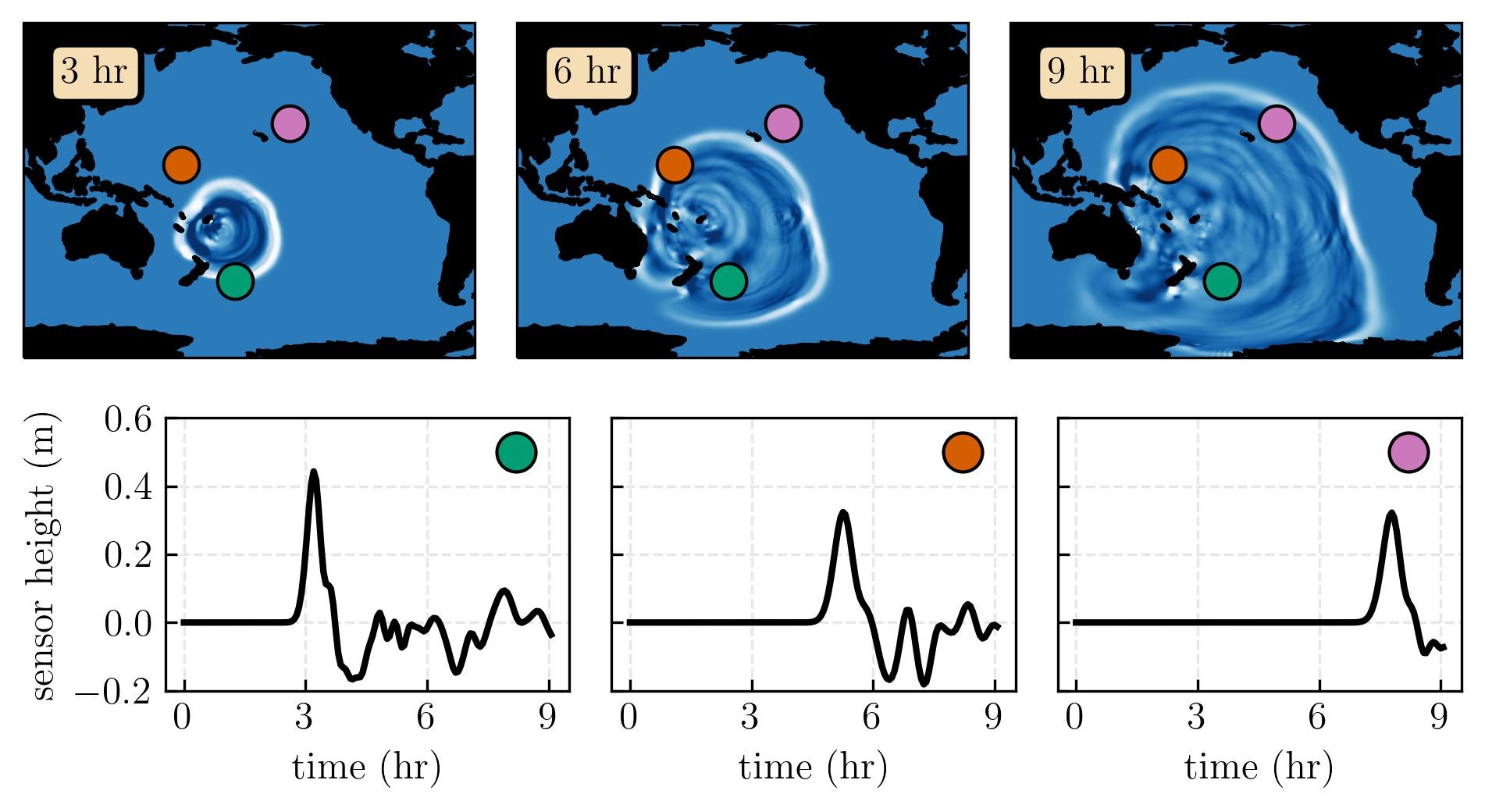}
\centering
\vspace{-0.5cm}
\caption{\textit{Frames of a tsunami simulation and sample sensor readings}. Increments of every three hours of a tsunami simulation, emanating just south of Fiji with a source at 174$^\circ$E, 21$^\circ$S, at an approximated magnitude of 7. By 10-12 hours, the primary tsunami wave will have reached every Pacific coast, driving waves often exceeding 1 m in height. In this example, artificial buoy sensors (shown as colored dots) record the passing tsunami wave height.}
\label{fig:frames}
\end{figure}

\section{Methods}
\subsection{Simulating tsunamis with the shallow water equations}
\label{sec:sim}
The shallow water equations (SWE) describe the motion of a relatively thin hydrostatic layer of fluid with constant density. This thin layer is justified for many geophysical and atmospheric flows occurring in a shallow layer where the dynamical evolution on a horizontal length scale is many orders larger than the vertical length scale \cite{vallis2019essentials}, i.e. when horizontal motion drives the dynamics. In modeling oceanic flows, i.e. tsunami evolution, these equations also enforce a free-boundary condition at the surface (the topmost air/water interface) and a no-slip boundary condition at the ocean floor (bathymetry). Furthermore, interaction with land above the ocean surface (topography) is required to accurately model the feedback present from waves advancing and receding the coasts. 

Practically, the shallow-water equations (SWE) are a depth-averaged simplification of the three-dimensional Navier--Stokes equations, obtained under the shallow-layer approximation by averaging the horizontal velocity over the fluid depth and neglecting vertical structure. A Coriolis term is included to account for the Earth's rotation, and a drag term may be included to model dissipation at the ocean floor. For the derivation of the GP-ROM, we write the SWE as a nonlinear PDE system governing the vector-valued state \(\vb q(\vb x,t)\) on the spatial domain \(\vb x\in\Omega\):
\begin{equation}
\pdv{\vb q}{t} + \mathcal{L}(\vb q) + \mathcal{N}(\vb q,\vb q) = 0.
\label{eq:pde}
\end{equation}

Here, \(\mathcal{L}\) and \(\mathcal{N}\) denote the linear and bilinear operators, respectively. In the present formulation, the state is written in terms of the fluid thickness and depth-averaged horizontal velocities,
\begin{equation}
\vb q
=
\begin{pmatrix}
u\\
v\\
h
\end{pmatrix},
\end{equation}
with
\begin{equation}
\mathcal{L}(\vb q)
=
\begin{pmatrix}
-g h_x + f v\\
-g h_y - f u\\
0
\end{pmatrix},
\qquad
\mathcal{N}(\vb q_1,\vb q_2)
=
\begin{pmatrix}
u_1 (u_2)_x + v_1 (u_2)_y\\
u_1 (v_2)_x + v_1 (v_2)_y\\
u_1 (h_2)_x + h_1 (u_2)_x + v_1 (h_2)_y + h_1 (v_2)_y
\end{pmatrix}.
\end{equation}

Here, \(h\) is the fluid thickness, \(u\) and \(v\) are the depth-averaged horizontal velocities in the \(x\)- and \(y\)-directions, \(f\) is the Coriolis parameter, and \(g\) is the gravitational acceleration. We distinguish the thickness \(h\) from the free-surface elevation \(\eta\), which satisfies
\begin{equation*}
\eta = h + z_b,
\end{equation*}
where \(z_b\) denotes the prescribed bathymetry, measured relative to the reference surface \(z=0\) and negative below this surface. In the present reduced-order formulation, \(h\) is treated as the prognostic height variable. Subscripts denote partial derivatives with respect to the corresponding spatial coordinate.

We note explicitly that the drag term is omitted from the reduced-order derivation, although it is included in the full SWE simulations. While this term is important for accurately resolving tsunami dynamics, it introduces a non-polynomial nonlinearity that cannot be directly incorporated into the precomputed GP-ROM operators. As discussed in Sec. \ref{sec:gprom}, we find that omitting this term in the operator construction does not significantly compromise model fidelity.

To simulate tsunamis, we use the swe-python code \cite{swegithub, mcdugald2025attention}, which solves a conservative form of these equations. There are two options that we employ to study earthquake-induced tsunamis: the first is via realistic conditions, provided by in-depth fault analysis that yield complex initial conditions interpolated to the mesh; the second is via a simple Gaussian displacement, wherein a Gaussian volume of water is lifted above the surface at the epicenter and released at the initial timestep. In the first case we require a detailed analysis of initial conditions by solving the dynamics of the full underwater earthquake event via GEOCLAW \cite{berger2011geoclaw}. And in the second, the initial conditions to the solver are an epicenter longitude, latitude, and magnitude scaling factor which controls the shape of the Gaussian volume -- this last simplification is to emulate earthquakes of varying magnitude, though at present no direct mapping between the Gaussian initial condition and a real earthquake magnitude exists. 

Currently, ocean buoy sensors such as the Deep-ocean Assessment and Reporting of Tsunami Network (DART) network of buoys are among the most reliable measurement tools of ocean wave height, including those resulting from seismic events, weather conditions, and tsunami waves \cite{gonzalez1998deep}. We emulate the buoy diagnostic process in our simulations, employing both synthetic buoy locations and using real-world DART buoy locations. Fig. \ref{fig:frames} demonstrates some simulation frames, contoured to highlight the intricate patterns of the tsunami evolution, for a tsunami emanating southeast of Fiji. The initial conditions to this tsunami are 174$^\circ$E, 21$^\circ$S, and a magnitude scaling factor of 2. In the second row, artificial sensors measure the height of the tsunami wave. For all subsequent analysis until the results of Section \ref{sec:results_fiji}, we use this simulation as a reference for derivations.

\subsection{Generating the modes and coefficients via Proper Orthogonal Decomposition}
\label{sec:pod}

The first step in the reduced-order modeling procedure is to construct a single, global POD basis for the full shallow-water state. At each snapshot time
\(t_\ell\), the discrete state is written as
\begin{equation*}
\vb q(t_\ell)
=
\begin{bmatrix}
\vb u(t_\ell)\\
\vb v(t_\ell)\\
\vb h(t_\ell)
\end{bmatrix}
\in \mathbb R^{3N},
\end{equation*}

where \(N\) is the number of spatial degrees of freedom for each state
component. We then form the snapshot matrix by arranging the discrete state
vectors as columns,
\begin{equation*}
\vb X
=
\begin{bmatrix}
\vb q(t_0) & \vb q(t_1) & \cdots & \vb q(t_{n-1})
\end{bmatrix}
\in \mathbb R^{3N\times n},
\end{equation*}
where \(n\) denotes the number of snapshots.

Next, we subtract the temporal mean state,
\begin{equation*}
\bar{\vb q}
=
\frac{1}{n}\sum_{\ell=0}^{n-1}\vb q(t_\ell),
\end{equation*}

to obtain the fluctuation snapshot matrix
\begin{equation*}
\vb Q
=
\begin{bmatrix}
\vb q(t_0)-\bar{\vb q} &
\vb q(t_1)-\bar{\vb q} &
\cdots &
\vb q(t_{n-1})-\bar{\vb q}
\end{bmatrix}
\in \mathbb R^{3N\times n}.
\end{equation*}

This mean subtraction centers the data about the temporal average and allows the POD modes to represent the dominant fluctuating dynamics, while the mean state is retained separately in the reduced reconstruction \cite{rowley2004model}.

We then compute the singular value decomposition (SVD) of \(\vb Q\),
\begin{equation}
\vb Q = \boldsymbol{\Phi}\boldsymbol{\Sigma}\vb V^{T},
\label{eq:svd}
\end{equation}

where
\begin{equation*}
\boldsymbol{\Phi}\in\mathbb R^{3N\times r},\qquad
\boldsymbol{\Sigma}\in\mathbb R^{r\times r},\qquad
\vb V\in\mathbb R^{n\times r},
\end{equation*}

and \(r=\operatorname{rank}(\vb Q)\). The columns of \(\boldsymbol{\Phi}\) are the POD modes and form an orthonormal basis for the spatial structure of the fluctuation field. The present work adopts the vector-valued POD inner product on the concatenated state \((u,v,h)\). While alternative dimensionally consistent or physics-based inner products are possible and may improve robustness, the joint \(L^2\)-type construction used here remains a common reduced-order modeling choice \cite{parish2023impact}; see also the general POD/Galerkin ROM framework reviewed by \cite{rowley2017model} and related POD-based reduced-order formulations for the shallow-water equations in \cite{ahmed2020sampling}.

The corresponding temporal coefficients are obtained by projection onto the POD
basis,
\begin{equation*}
\vb A
=
\boldsymbol{\Phi}^{T}\vb Q
\in \mathbb R^{r\times n}.
\end{equation*}
The \(\ell\)-th column of \(\vb A\) is therefore the coefficient vector
associated with snapshot \(t_\ell\),
\begin{equation*}
\vb a(t_\ell)
=
\begin{bmatrix}
a_1(t_\ell) & \cdots & a_r(t_\ell)
\end{bmatrix}^{T}
\in \mathbb R^{r}.
\end{equation*}

Since the POD is performed on the concatenated shallow-water state, each mode
contains contributions from all state variables,
\begin{equation*}
\boldsymbol{\phi}_i
=
\begin{bmatrix}
\boldsymbol{\phi}_i^u\\
\boldsymbol{\phi}_i^v\\
\boldsymbol{\phi}_i^h
\end{bmatrix},
\qquad i=1,\dots,r.
\end{equation*}
Thus, each scalar coefficient \(a_i(t)\) multiplies the entire modal vector
\(\boldsymbol{\phi}_i\), including its \(u\)-, \(v\)-, and \(h\)-components.
Consequently, the reduced representation employs a single shared coefficient
vector \(\vb a(t)\), rather than separate coefficient vectors for the
individual state variables.

Using the complete POD basis, the full state can be reconstructed exactly
as
\begin{equation*}
\vb q(\vx,t)
=
\bar{\vb q}(\vx)
+
\sum_{i=1}^{r}
a_i(t)\boldsymbol{\phi}_i(\vx).
\end{equation*}

Expressed componentwise, this yields
\begin{equation*}
\vb u(\vx,t)
=
\bar{\vb u}(\vx)
+
\sum_{i=1}^{r}
a_i(t)\bs{\phi}_i^u(\vx),
\end{equation*}
\begin{equation*}
\vb v(\vx,t)
=
\bar{\vb v}(\vx)
+
\sum_{i=1}^{r}
a_i(t)\bs\phi_i^v(\vx),
\end{equation*}
\begin{equation*}
\vb h(\vx,t)
=
\bar{\vb h}(\vx)
+
\sum_{i=1}^{r}
a_i(t)\bs\phi_i^h(\vx).
\end{equation*}

In practice, only a subset of the POD modes is retained to obtain a
reduced-order representation; this truncation is discussed in the following
section.

\begin{figure}[h!]
\includegraphics[width=\textwidth]{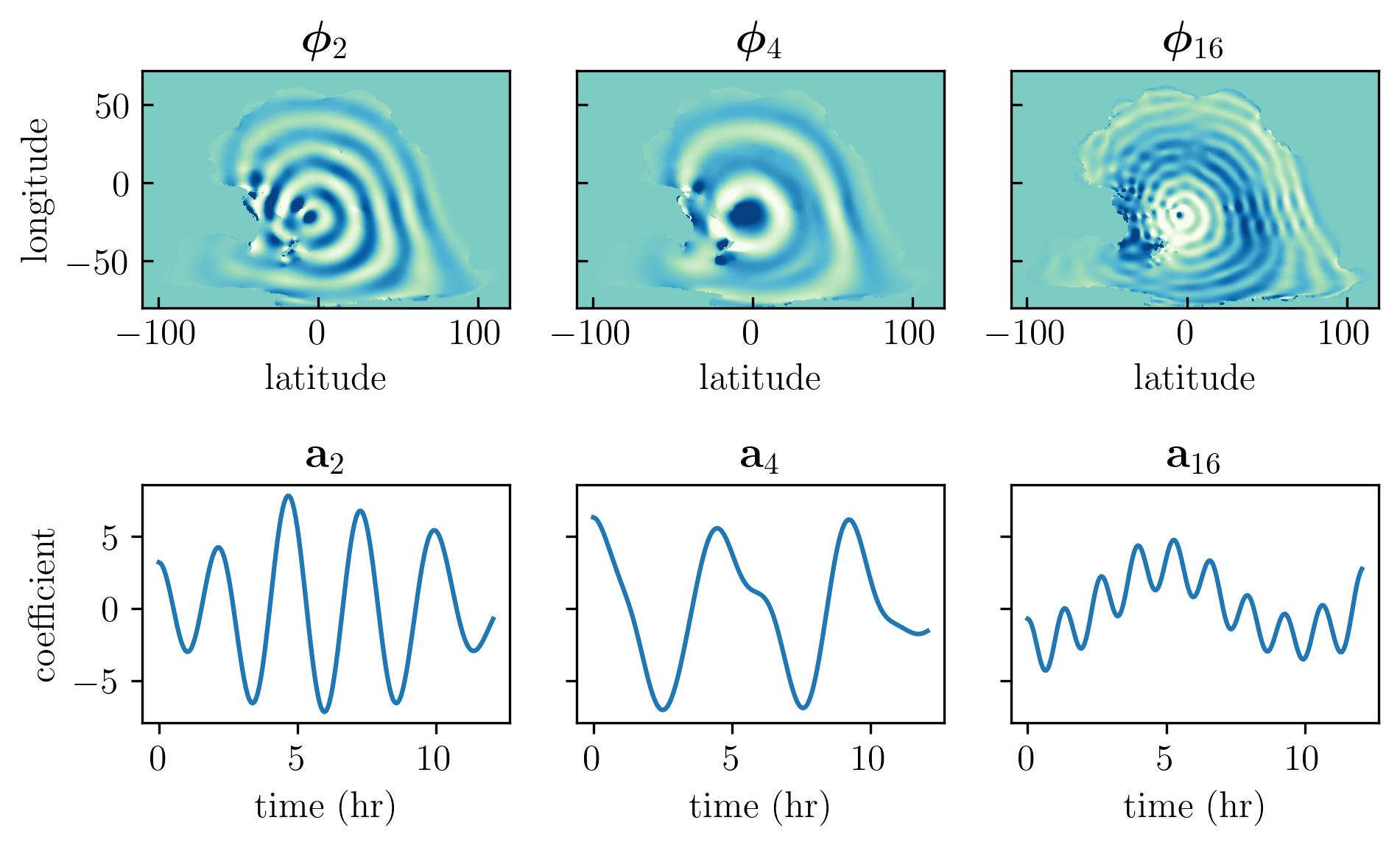}
\centering
\vspace{-0.5cm}
\caption{\textit{Modes and coefficients of the Fiji tsunami simulation}. Selected spatial modes $\bm{\phi}_i$ of the reference simulation shown in Figure \ref{fig:frames} (we show only the height component of these modal vectors.) In the second row, the corresponding temporal coefficients show the generally observed quasi-periodic time evolution of these modes.}
\label{fig:modes}
\end{figure}

In Fig. \ref{fig:modes} we demonstrate these spatial modes and temporal coefficients for the simulation snapshots shown in Fig. \ref{fig:frames}. We note that in this figure, as well as all figures in the paper where spatial information is used, we only employ the height component of the mode. However, as discussed later, the entire modal vector (i.e. $h$, $u$, $v$ components) is used for construction of the GP-ROM.

\subsubsection{Mode truncation}
To obtain a reduced-order model from POD, we retain only the leading \(k\) modes, where typically \(k \ll r\). This mode truncation yields the approximate reconstruction

\begin{equation}
\vb q(\vx,t)
\approx
\bar{\vb q}(\vx)
+
\sum_{i=1}^{k}\boldsymbol{\phi}_i(\vx)a_i(t).
\label{eq:approx_expansion}
\end{equation}

The singular values in \(\boldsymbol{\Sigma}\) from Eq. \ref{eq:svd} provide a useful measure for determining how many modes are required to obtain an accurate low-rank approximation of the fluctuation field. We quantify this using the relative informational content (RIC), defined as

\begin{equation}
\mathrm{RIC}(m)
=
\frac{\sum_{i=1}^{m}\sigma_i}{\sum_{i=1}^{r}\sigma_i},
\end{equation}

where \(\sigma_i\) denotes the \(i\)-th singular value.

\begin{figure}[h!]
\includegraphics[width=0.8\textwidth]{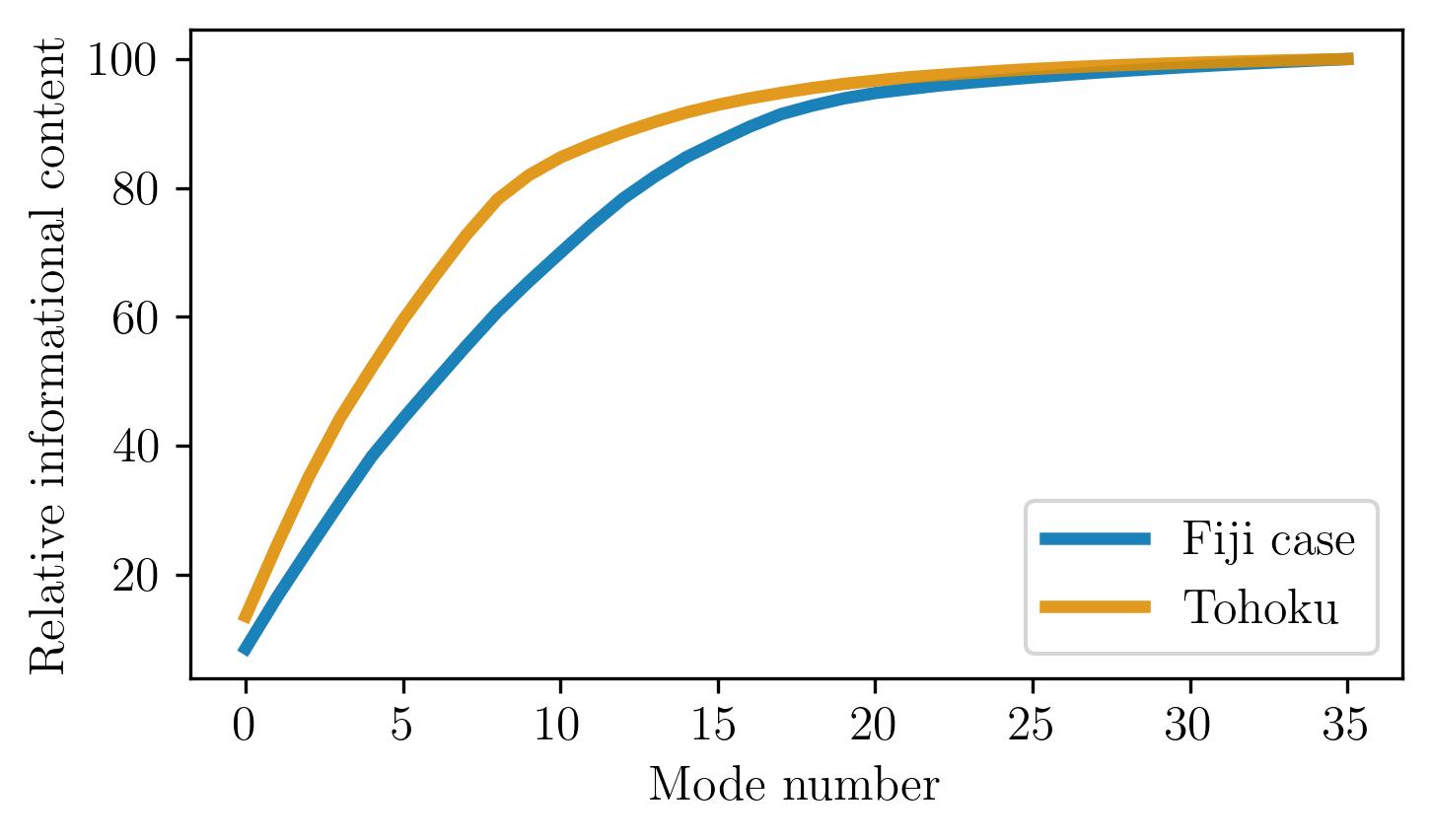}
\centering
\vspace{-0.5cm}
\caption{\textit{Relative informational content of the POD modes}. Retaining 16 modes captures roughly \(80\%\) of the total informational content, while retaining 24 modes captures approximately \(95\%\).}
\label{fig:ric}
\end{figure}

In the present simulations, approximately 200 snapshots are retained, yielding at most 200 POD modes. In practice, we retain approximately \(10\%\) of these modes. At this truncation level, the reconstruction error associated with Eq. \ref{eq:approx_expansion} is less than \(8\%\) in the \(L_2\) norm. Figure \ref{fig:ric} shows the RIC for the cases considered in this study and is used to guide the truncation point. For these cases, retaining roughly 20 modes captures approximately \(85\%\) to \(92\%\) of the total informational content. Figure \ref{fig:ric} includes four simulations: three from the Fiji case study described in Sec. \ref{sec:results_fiji} and one from the Tohoku case study described in Sec. \ref{sec:tohoku}.

\subsection{Constructing the GP-ROM}
\label{sec:gprom}

With the shallow-water equations cast in the operator form of Eq. \ref{eq:pde}, we substitute the truncated POD expansion of Eq. \ref{eq:approx_expansion},
\begin{equation}
\vb q(\vx,t)
\approx
\bar{\vb q}(\vx)
+
\sum_{i=1}^{k}\boldsymbol{\phi}_i(\vx)a_i(t),
\end{equation}
and then perform a Galerkin projection onto the POD basis. For each test function
\(\boldsymbol{\phi}_m\), \(m=1,\dots,k\), this yields
\begin{align}
\left\langle
\frac{\partial}{\partial t}
\left(
\bar{\vb q}
+
\sum_{i=1}^{k}a_i\boldsymbol{\phi}_i
\right),
\boldsymbol{\phi}_m
\right\rangle
&+
\left\langle
\mathcal{L}
\left(
\bar{\vb q}
+
\sum_{i=1}^{k}a_i\boldsymbol{\phi}_i
\right),
\boldsymbol{\phi}_m
\right\rangle
\nonumber\\
&+
\left\langle
\mathcal{N}
\left(
\bar{\vb q}
+
\sum_{i=1}^{k}a_i\boldsymbol{\phi}_i,\,
\bar{\vb q}
+
\sum_{j=1}^{k}a_j\boldsymbol{\phi}_j
\right),
\boldsymbol{\phi}_m
\right\rangle
= 0.
\end{align}

Because the POD modes are orthonormal,
\begin{equation*}
\langle \boldsymbol{\phi}_i,\boldsymbol{\phi}_j\rangle = \delta_{ij},
\end{equation*}
where \(\delta_{ij}=1\) if \(i=j\) and \(\delta_{ij}=0\) otherwise, the projected equations may be grouped into constant (\(C\)), linear (\(L\)), and quadratic (\(Q\)) contributions. For the shallow-water equations written in the state variables \(\vb q=(u,v,h)\), the resulting operator coefficients are
\begin{align}
C_i &= \langle \phi_i^u, C^u \rangle + \langle \phi_i^v, C^v \rangle + \langle \phi_i^h, C^h \rangle, \\
L_{ij} &= \langle \phi_i^u, L_j^u \rangle + \langle \phi_i^v, L_j^v \rangle + \langle \phi_i^h, L_j^h \rangle, \\
Q_{ijm} &= \langle \phi_i^u, Q_{jm}^u \rangle + \langle \phi_i^v, Q_{jm}^v \rangle + \langle \phi_i^h, Q_{jm}^h \rangle.
\end{align}

The corresponding component fields are
\begin{align}
C^u &= -g\bar h_x + f\bar v - \bar u \bar u_x - \bar v \bar u_y, \\
C^v &= -g\bar h_y - f\bar u - \bar u \bar v_x - \bar v \bar v_y, \\
C^h &= -\bar u \bar h_x - \bar h \bar u_x - \bar v \bar h_y - \bar h \bar v_y, \\
L_j^u &= -g(\phi_j^h)_x + f\phi_j^v - \bar u(\phi_j^u)_x - \phi_j^u \bar u_x - \bar v(\phi_j^u)_y - \phi_j^v \bar u_y, \\
L_j^v &= -g(\phi_j^h)_y - f\phi_j^u - \bar u(\phi_j^v)_x - \phi_j^u \bar v_x - \bar v(\phi_j^v)_y - \phi_j^v \bar v_y, \\
L_j^h &= -\bar u(\phi_j^h)_x - \phi_j^u \bar h_x - \bar h(\phi_j^u)_x - \phi_j^h \bar u_x
        - \bar v(\phi_j^h)_y - \phi_j^v \bar h_y - \bar h(\phi_j^v)_y - \phi_j^h \bar v_y, \\
Q_{jm}^u &= -\phi_j^u(\phi_m^u)_x - \phi_j^v(\phi_m^u)_y, \\
Q_{jm}^v &= -\phi_j^u(\phi_m^v)_x - \phi_j^v(\phi_m^v)_y, \\
Q_{jm}^h &= -\phi_j^u(\phi_m^h)_x - \phi_j^h(\phi_m^u)_x - \phi_j^v(\phi_m^h)_y - \phi_j^h(\phi_m^v)_y.
\label{eq:coef_term_end}
\end{align}

These coefficients are assembled offline prior to time integration of the reduced system. The resulting reduced operators have dimensions \(\vb C\in\mathbb R^{k}\), \(\vb L\in\mathbb R^{k\times k}\), and \(\vb Q\in\mathbb R^{k\times k\times k}\). For systems with a large number of retained modes, their evaluation may be computationally expensive; in particular, the quadratic tensor requires \(O(k^3)\) coefficient evaluations and storage of \(O(k^3)\) entries. We also reiterate that the drag term is excluded from this derivation because it introduces a non-polynomial nonlinearity and therefore cannot be directly represented within the precomputed coefficient operators. As discussed in the following section, the corrected GP-ROM formulation is expected to account for unresolved dynamics introduced by this approximation.

The resulting reduced-order system is
\begin{equation}
\dot a_i
=
C_i
+
\sum_{j=1}^{k}L_{ij}a_j
+
\sum_{j=1}^{k}\sum_{m=1}^{k}Q_{ijm}a_j a_m.
\label{eq:ode_variable}
\end{equation}

Using Einstein summation notation, this may be written compactly as
\begin{equation}
\dot a_i = C_i + L_{ij}a_j + Q_{ijm}a_j a_m.
\label{eq:gprom}
\end{equation}

We refer to Eq. \ref{eq:gprom}, together with its associated initial conditions and parameters, as the GP-ROM. For a given simulation and its corresponding POD modes, the operators \(\vb C\), \(\vb L\), and \(\vb Q\) are computed once and the reduced system is then integrated using a standard ODE solver, such as a fourth-order Runge--Kutta method, with prescribed initial coefficients \(\vb a_0 \equiv \vb a(t=0)\). The GP-ROM therefore defines a mapping from the initial modal coefficients to the reduced trajectory, i.e.
\begin{equation*}
f:\vb a_0 \mapsto \vb a(t).
\end{equation*}

\begin{figure}[h!]
\includegraphics[width=0.9\textwidth]{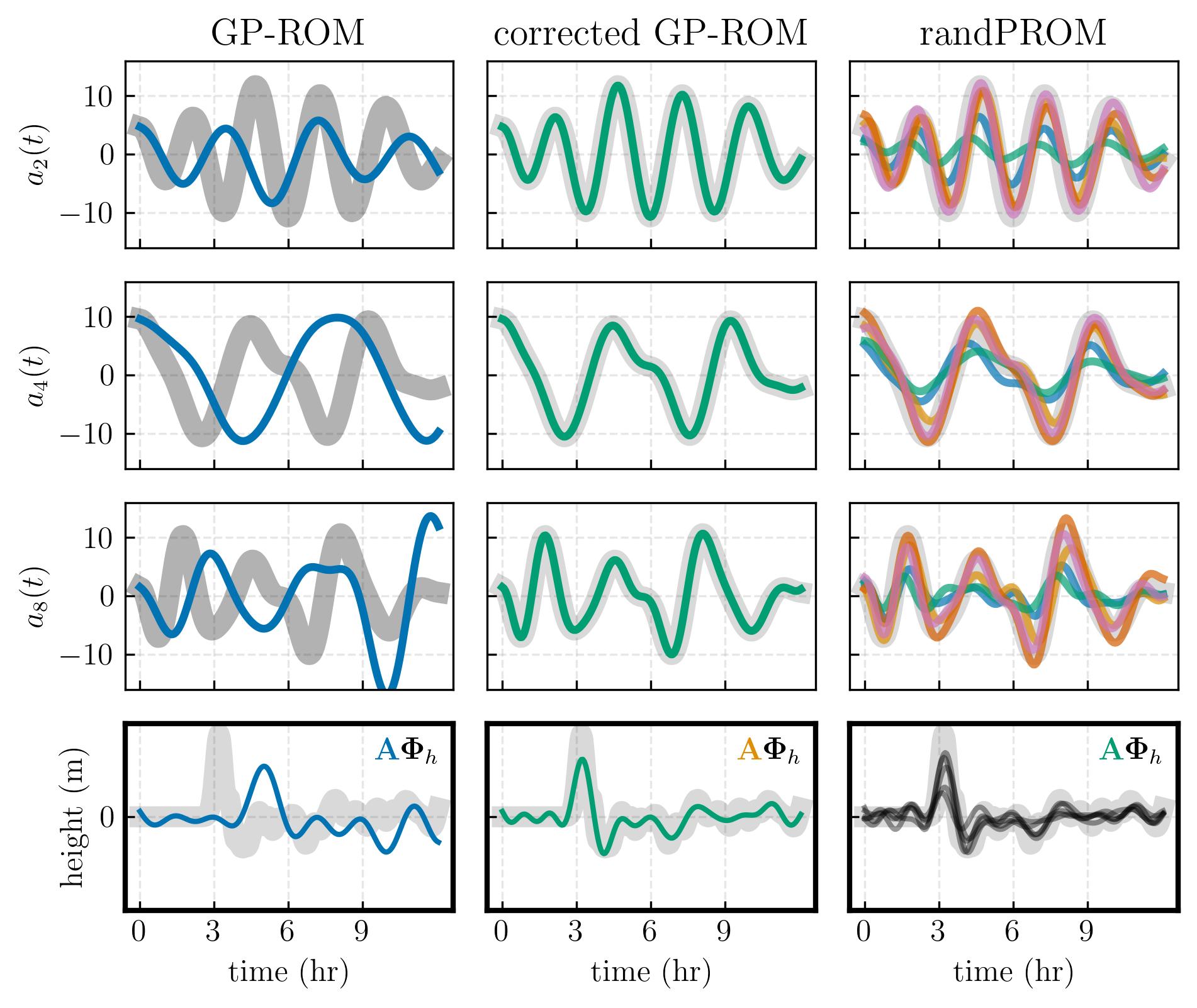}
\centering
\vspace{-0.5cm}
\caption{\textit{Creating new activations.} The first column shows the GP-ROM with its inherent instability, the second shows significantly improved stability after the operator-correction of the GP-ROM, and the third shows the randPROM, employing 8 posterior samples drawn from a learned distribution. The first three rows show activations $a_2$, $a_4$, and $a_{16}$ (in color), respectively, against the true SVD activations (in gray). The final row shows an example reconstruction of a sensor, a key result. }
\label{fig:coefs_all}
\end{figure}

For the Fiji simulation (Fig. \ref{fig:frames}), we construct the GP-ROM using the theory described above. When these coefficients are recombined with the POD modes, the resulting reconstruction is expected to approximate Eq. \ref{eq:approx_expansion}. In practice, however, the GP-ROM trajectories accumulate reduced-model error over time. The first column of Fig. \ref{fig:coefs_all} shows that the coefficients generated by the GP-ROM (blue lines) deviate from the POD coefficients (thick gray lines), primarily through long-time phase and amplitude mismatch. These discrepancies are not solely a consequence of dynamical instability, but also reflect truncation, projection, and basis-representation error. More generally, although a POD basis may be constructed from a broader class of scenarios, the resulting reduced model remains constrained by the span of the retained modes and by the data used to construct them. These errors are further amplified in the sensor-height reconstruction shown in the bottom row of the same column.

These issues are remediated in several ways, the most accessible being pre-calibration of the operators $\vb C$, $\vb L$, and $\vb Q$. For example, we scale the operators $\vb C$, $\vb L$, and $\vb Q$ by constants 0.008, 1.036,  and 1.01, respectively. This constant scaling is a known remediation, found iteratively, that is in practice imprecise and often cannot fix other structural issues. 

We fix this behavior further via operator-correction as discussed in the next section, and explain how stabilizing the GP-ROM produces the significantly improved coefficients, as shown in the following middle column of Figure \ref{fig:coefs_all}.

\begin{figure}[t!]
\includegraphics[width=\textwidth]{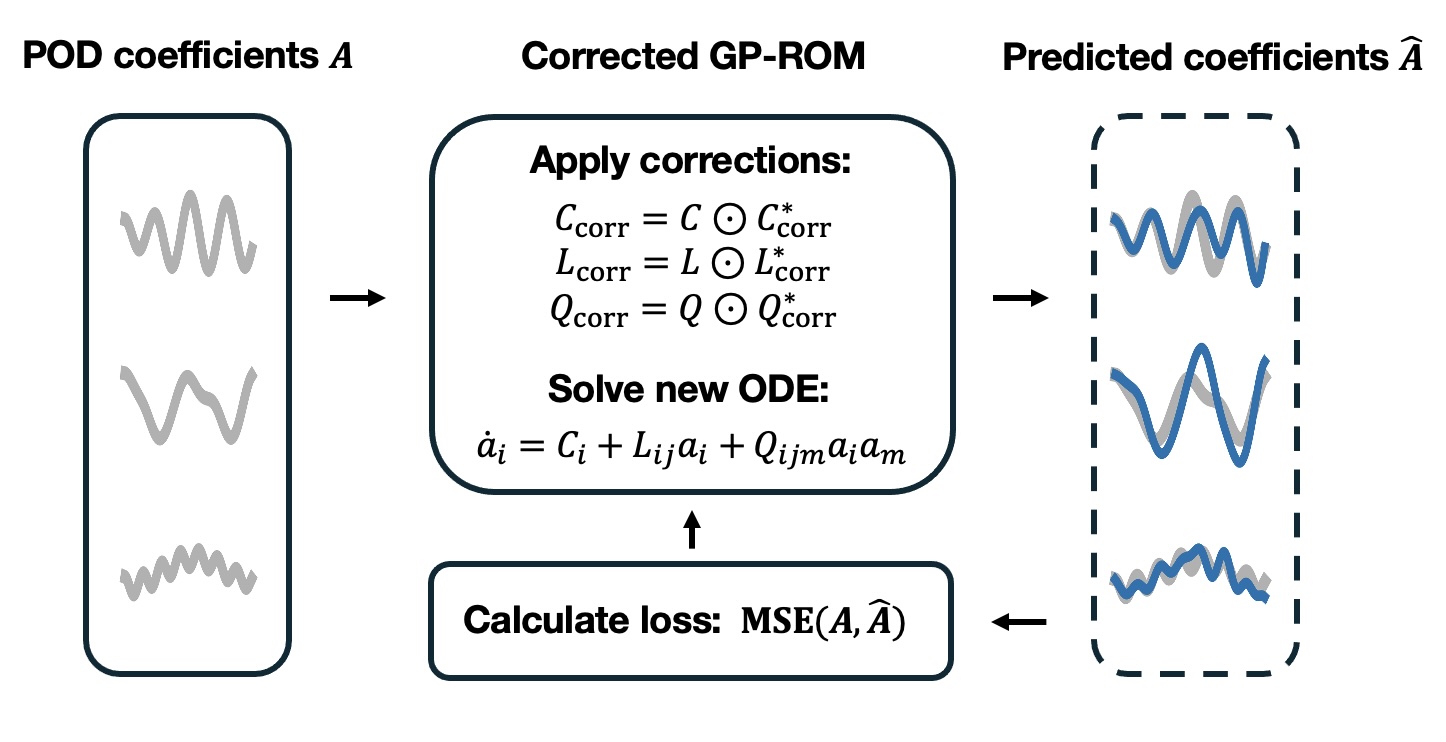}
\centering
\vspace{-0.5cm}
\caption{\textit{Training the operator-corrected GP-ROM}. The input to the corrected GP-ROM is the POD coefficient trajectory. At each training iteration, the method updates corrections to the corrected, linear, and quadratic coefficient matrices $\vb C$, $\vb L$, and $\vb Q$, respectively, and then solves the corresponding reduced ODE.}
\label{fig:cgp}
\end{figure}

\subsection{Operator optimization to produce the corrected GP-ROM}

Because the trajectories of the GP-ROM coefficients may be unstable or inaccurate, we introduce a calibrated GP-ROM that corrects the reduced operators for a given initial condition. Similar to the GP-ROM, this corrected model defines a mapping from the initial modal coefficients to the full reduced trajectory, i.e.
\begin{equation*}
g:\vb a_0 \mapsto \vb a(t).
\end{equation*}

The calibration is performed by learning elementwise correction factors applied to the reduced operators. In the present formulation, the corrected operators are written as
\begin{equation}
    \label{eq:cgp_op}
    \vb C_{\mathrm{corr}} = \vb C \odot \vb C_{\mathrm{corr}}^{*},
    \hspace{0.5cm}
    \vb L_{\mathrm{corr}} = \vb L \odot \vb L_{\mathrm{corr}}^{*},
    \hspace{0.5cm}
    \vb Q_{\mathrm{corr}} = \vb Q \odot \vb Q_{\mathrm{corr}}^{*}.
\end{equation}
For each training iteration, the method updates these correction factors, solves the corresponding reduced ODE, and evaluates the resulting trajectory against the reference POD coefficients using an \(L_2\)-based loss, together with a regularization term that promotes long-time stability.

After training, the calibrated GP-ROM produces activation trajectories that are both accurate and stable (second column of Fig. \ref{fig:coefs_all}). The output of this procedure is the set of corrected operators in Eq. \ref{eq:cgp_op}, which are then used in the subsequent inference stage. In addition to accurately reproducing the activation trajectories, reconstructions of the spatiotemporal PDE solution also remain highly accurate, as shown in the last row of the second column of Fig. \ref{fig:coefs_all}.

However, the corrected GP-ROM remains limited to the reduced representation identified by the POD analysis. In the present study, our primary interest is not in extending the time integration beyond the available training data, but in using the stable corrected reduced model as the forward model for inference. In the following section, we therefore learn distributions over the initial modal coefficients from sensor observations so that we may infer event-specific reduced trajectories for new tsunami scenarios.

\subsection{Calibrating randPROMs to sparse observational data via Bayesian hierarchical pooling}
\label{sec:bayesian}

To motivate the need for a more predictive reduced model, consider the Fiji tsunami simulation with state \(\vb q\). From this simulation, we obtain POD modes \(\boldsymbol{\Phi}\) and corresponding coefficients \(\vb A\), such that
\begin{equation*}
\vb q \approx \boldsymbol{\Phi}\vb A.
\end{equation*}

Now suppose that we observe sparse sensor data from a related tsunami event, or alternatively consider a new tsunami simulation with different initial conditions. Let the corresponding state of this new scenario be denoted by \(\vb y\). If \(\vb y\) lies approximately in the subspace spanned by \(\boldsymbol{\Phi}\), then it may be represented as
\begin{equation*}
\vb y \approx \boldsymbol{\Phi}\vb B,
\end{equation*}
where \(\vb B\) denotes the coefficients of the new state in the POD basis \(\boldsymbol{\Phi}\).

If the full state \(\vb y\) were known, these coefficients could be obtained by solving the least-squares problem
\begin{equation}
\vb B = \arg\min_{\vb B}\norm{\boldsymbol{\Phi}\vb B - \vb y}_2.
\label{eq:lsq}
\end{equation}
Thus, while \(\vb A\) corresponds to the coefficients of the original Fiji simulation, \(\vb B\) corresponds to the coefficients of a new state projected onto the same reduced basis.

However, this formulation has several important limitations. First, if \(\vb y\) corresponds to a real observed event, then the state is not fully known in either space or time. In practice, only sparse and noisy measurements may be available from a limited set of sensors, making Eq. \ref{eq:lsq} underdetermined. Second, the reconstruction depends on the sensor configuration and observation locations, so the solution of Eq. \ref{eq:lsq} is inherently tied to the observation operator associated with a given sensor network. Third, Eq. \ref{eq:lsq} alone does not provide a natural mechanism for quantifying uncertainty or for borrowing information across related scenarios. These limitations motivate a probabilistic calibration framework.

Our approach is to regularize this inverse problem within a Bayesian hierarchical model. Suppose now that we have \(i=1,\ldots,n\) tsunami simulations or observational cases, which we refer to as calibration scenarios and denote by \(\vb y_i\). Let \(\boldsymbol{\Phi}\) be a fixed set of POD modes, and let \(g(\cdot,t)\) denote the corrected reduced-order evolution operator constructed on this basis. We model the calibration scenarios using
\begin{align}
\vb y_i(t) &= \bm H_i(t)\boldsymbol{\Phi}\vb B_i(t) + \bm\varepsilon_i(t), \label{eq:goal} \\
\vb B_i(t) &= g(\vb b_{0i}, t), \label{eq:cgp} \\
\vb b_{0i} &\sim \mathcal N(\bm\mu_0,\bm\Sigma), \label{eq:init} \\
\bm\varepsilon_i(t) &\sim \mathcal N(\bm 0,\vb\Sigma_\varepsilon). \label{eq:noise}
\end{align}

Here, \(\vb B_i(t)\) denotes the reduced coefficients for scenario \(i\), generated by the corrected reduced model from the latent initial coefficient vector \(\vb b_{0i}\). The observation matrix \(\bm H_i(t)\) maps the reconstructed state to the available sensor measurements and allows for scenario-specific sensor locations and missing observations. The residual term \(\bm\varepsilon_i(t)\) accounts for observation error as well as model discrepancy.

In this hierarchical formulation, the latent initial coefficients
\(\{\vb b_{0i}\}_{i=1}^{n}\) are scenario-specific and therefore define the
local level of the model. By contrast, the Gaussian prior parameters
\(\bm\mu_0\) and \(\bm\Sigma\) are shared across all scenarios and therefore define the global level of the hierarchy. Thus, “local” refers to inference for an individual observed event, whereas “global” refers to the population-level distribution from which event-specific initial conditions are assumed to arise.

Under this formulation, the latent initial conditions \(\{\vb b_{0i}\}_{i=1}^{n}\) are linked through the shared Gaussian prior in Eq. \ref{eq:init}, which induces hierarchical pooling across the calibration scenarios. This allows information from multiple observed events to regularize the inference of scenario-specific initial coefficients. Posterior inference is performed using a Markov chain Monte Carlo (MCMC) algorithm, specifically an adaptive Metropolis--Hastings method within a Gibbs sampler, in which the scenario-specific latent initial conditions \(\{\vb b_{0i}\}_{i=1}^{n}\) are updated with adaptive Metropolis--Hastings steps and the remaining hierarchical parameters are updated through conjugate Gibbs steps. In our experiments, we run 5000 burn-in iterations followed by 5000 sampling iterations.

The structure in Eqs. \eqref{eq:goal}--\eqref{eq:noise} therefore provides a framework for learning the posterior distribution of the latent initial values \(\{\vb b_{0i}\}_{i=1}^{n}\) from sparse observations. In particular, the observation-error variance may absorb approximation error in the corrected reduced model, uncertainty arising from differences between the calibration scenario and the reference state used to construct the basis, and measurement noise at the sensors. In this way, the model produces probabilistically calibrated reconstructions that remain consistent with the reduced dynamics.

Importantly, the formulation in Eqs. \eqref{eq:goal}--\eqref{eq:noise} should be interpreted as a calibration model rather than a fully predictive emulator for unseen events. Because it does not condition on external scenario descriptors, such as earthquake location or magnitude, it does not by itself distinguish new events in the absence of scenario-specific observations. Prediction for new events without direct calibration data therefore requires additional modeling assumptions, which are introduced in the following sections. 

We note that the formulation in \Crefrange{eq:goal}{eq:noise} shares similarities with data assimilation (DA) frameworks, in that observations are combined with a dynamical model to infer latent states. Both DA and randPROM frameworks can incorporate new observational data; however, in the randPROM framework this is achieved through fine-tuning the model rather than performing sequential state estimation. In contrast to many DA approaches which may rely on a fixed forecast model, we employ a operator-corrected GP-ROM (cGP) generated from a reference scenario. Furthermore, whereas DA primarily aims to estimate the best current state, the randPROM framework is designed to construct a reduced-order model with uncertainty quantification.

\subsubsection{Example of the hierarchical structure}
\begin{figure}[t!]
\centering 
\includegraphics[width=0.8\textwidth]{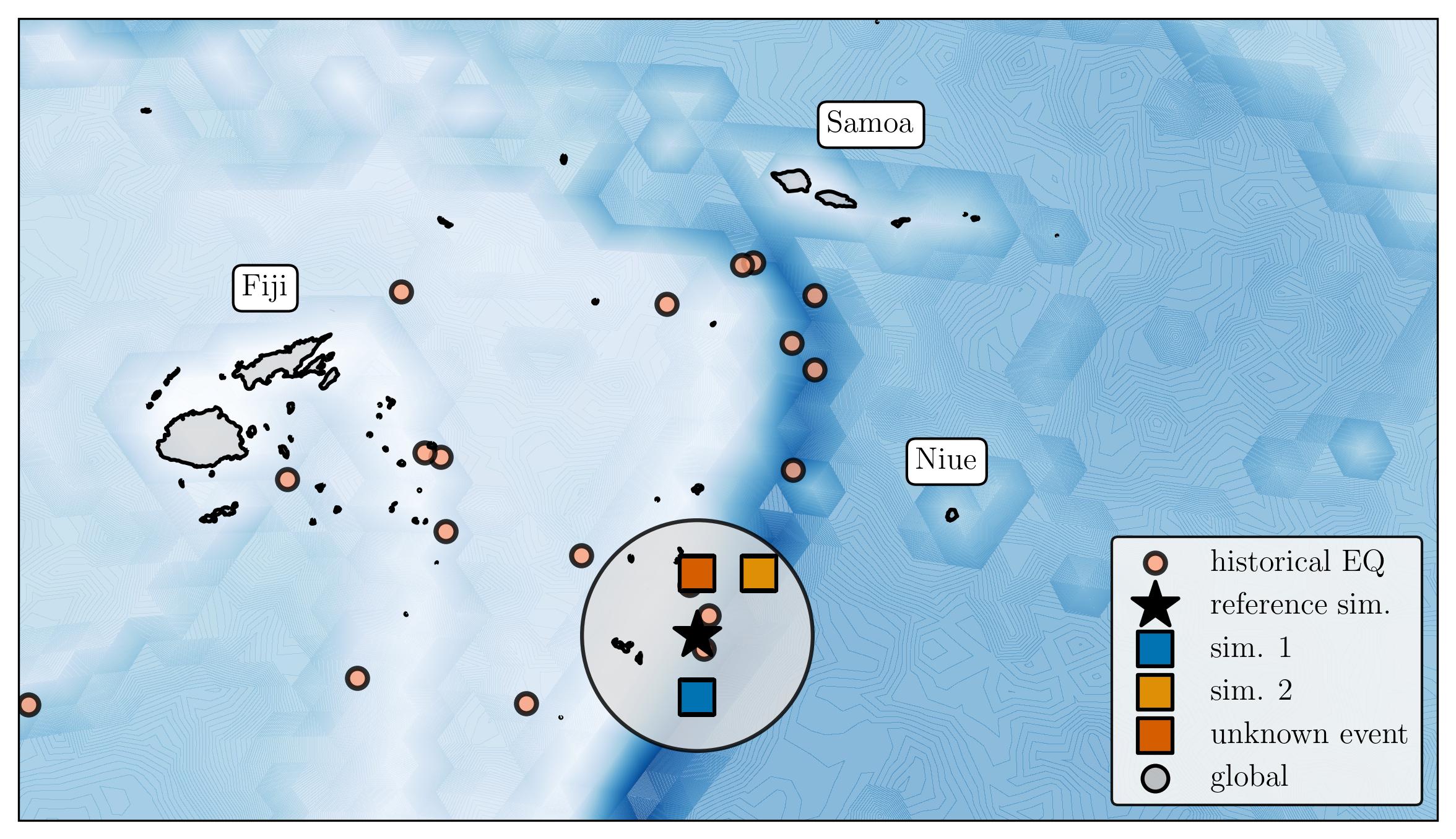}
\caption{\textit{Illustration of the hierarchical calibration structure for a region near Samoa.} A reduced-order model is constructed from a reference simulation, and additional simulations in the same event family are used as calibration scenarios in Eqs. \ref{eq:goal}--\ref{eq:noise}. In the Bayesian hierarchy, each calibration scenario has its own local latent initial condition, while the corresponding population distribution is described by shared global parameters. The epicenters of selected historical earthquakes are shown only to motivate the regional example.}
\label{fig:hierarchical_example}
\end{figure}

To illustrate the hierarchical model, we consider the schematic example shown in Fig. \ref{fig:hierarchical_example}. Suppose that we focus on a region of interest near Samoa, motivated by the locations and magnitudes of historical earthquake events in the broader Fiji domain. Within this region, we consider one simulation as a reference scenario and several additional simulations as calibration scenarios. The reference scenario is used to construct the reduced basis and corrected reduced-order dynamics, while the remaining simulations (sim. 1 and sim.2) are used to calibrate the Bayesian model in Eqs. \ref{eq:goal}--\ref{eq:noise}. Only one scenario (unknown event) is treated as an event that we attempt to explain with the calibrated model.

In this setting, the hierarchy is defined in parameter space rather than physical space. Each calibration scenario has its own latent initial reduced state \(\vb b_{0i}\), which forms the local level of the model. These scenario-specific latent states are linked through the shared population parameters \(\bm\mu_0\) and \(\bm\Sigma\), which form the global level of the hierarchy. Thus, the hierarchical structure reflects partial pooling across related scenarios within the chosen event family.

The purpose of this construction is to learn a population-level distribution for event-specific reduced initial conditions from multiple calibration scenarios. This provides a probabilistic calibration framework for related events in the region of interest. Prediction for a genuinely new event, however, requires either scenario-specific observations for calibration or additional assumptions introduced in the following section.

\subsection{A Bayesian hierarchical calibration model}
\label{sec:fullfull}

\begin{figure}[t!]
\centering 
\includegraphics[width=\textwidth]{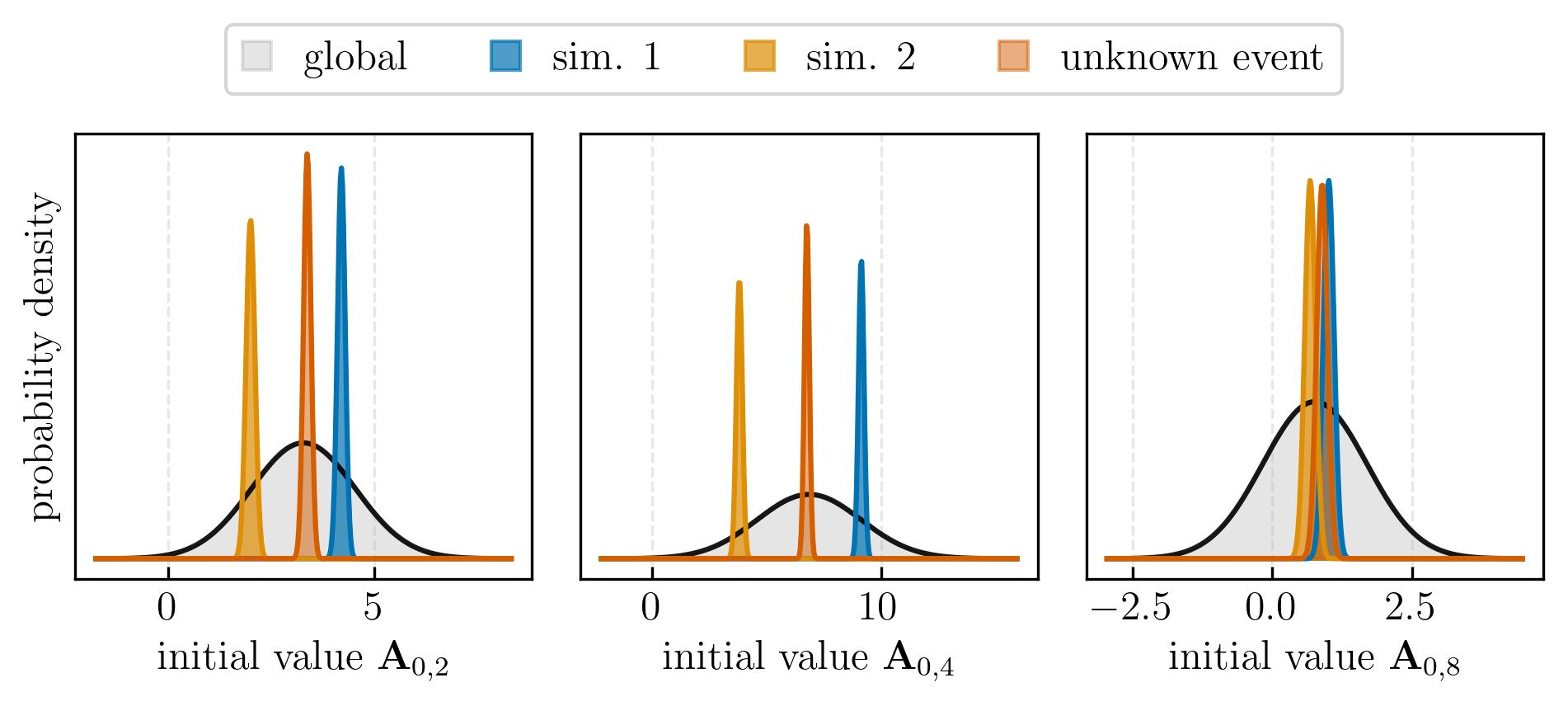}
\caption{\textit{Probability densities of selected latent initial values for the hierarchical example of Fig. \ref{fig:hierarchical_example}.} Each calibration scenario has its own posterior distribution for the local initial reduced state. These scenario-specific distributions are linked through a shared global distribution over initial reduced conditions, which captures population-level variation across the calibration scenarios.}
\label{fig:hierarchical_densities}
\end{figure}

The primary goal is to learn a distribution over the latent initial reduced coefficients \(\vb b_{0i}\) such that the resulting posterior predictive distribution is both accurate and calibrated. For the first randPROM demonstration, we use the hierarchical model in Eqs. \ref{eq:goal}--\ref{eq:noise} together with the priors
\begin{equation}
    \bm{\mu}_0 \sim \mathcal{N}(\vb A_0, \boldsymbol{\Sigma}_0), \qquad
    \vb{\Sigma} \sim \mathcal{W}^{-1}(\Psi, \nu), \label{eq:fullfull}
\end{equation}
where \(\mathcal{W}^{-1}(\Psi,\nu)\) denotes an inverse-Wishart distribution with scale matrix \(\Psi\) and degrees of freedom \(\nu\). Here, \(\bm\mu_0\) is the population-level mean of the latent initial reduced states, while \(\vb\Sigma\) is their population covariance across calibration scenarios. We use the reference initial condition \(\vb A_0\) as the prior mean for \(\bm\mu_0\), thereby centering the hierarchy on the nominal reduced initial state of the reference simulation. The hyperparameters \(\boldsymbol{\Sigma}_0\), \(\Psi\), and \(\nu\) control the spread and regularization of the hierarchical prior. In particular, \(\boldsymbol{\Sigma}_0\) specifies the prior covariance for \(\bm\mu_0\), while \(\Psi\) and \(\nu\) determine the scale and concentration of the inverse-Wishart prior on \(\vb\Sigma\). In our implementation, these prior covariances are chosen to be diagonal, with a relatively tight prior assigned to the leading retained mode and broader priors assigned to the remaining modes. Figure \ref{fig:hierarchical_densities} provides a schematic view of this hierarchy in coefficient space for the Samoa example of Fig. \ref{fig:hierarchical_example}. Each calibration scenario contributes its own posterior distribution over the local latent initial condition, while the shared global distribution captures the broader event-family variability implied by all calibration scenarios together.

\section{Applying the randPROM architecture to simulations of tsunamis near Fiji}
\label{sec:results_fiji}

To demonstrate the calibration process of the randPROM model, we begin with a synthetic study based on the Fiji tsunami used previously throughout the derivations. The reference event has source location \(174^\circ\mathrm{E},\,21^\circ\mathrm{S}\) and magnitude scaling factor 2. To generate the calibration scenarios \(\vb y_i\), we perform a suite of simulations obtained by perturbing the reference initial condition. These perturbation simulations use epicenter longitude \(174^\circ\mathrm{E}\pm1^\circ\), epicenter latitude \(21^\circ\mathrm{S}\pm1^\circ\), and magnitude scaling factor \(2\pm1\). This yields 27 perturbation simulations, of which we retain the 15 most extreme cases, measured by minimum and maximum wave activity. Here, wave activity is defined as the percentage of time that a sensor records a nontrivial wave height.

\begin{figure}[t!]
\centering
\includegraphics[width=0.9\textwidth]{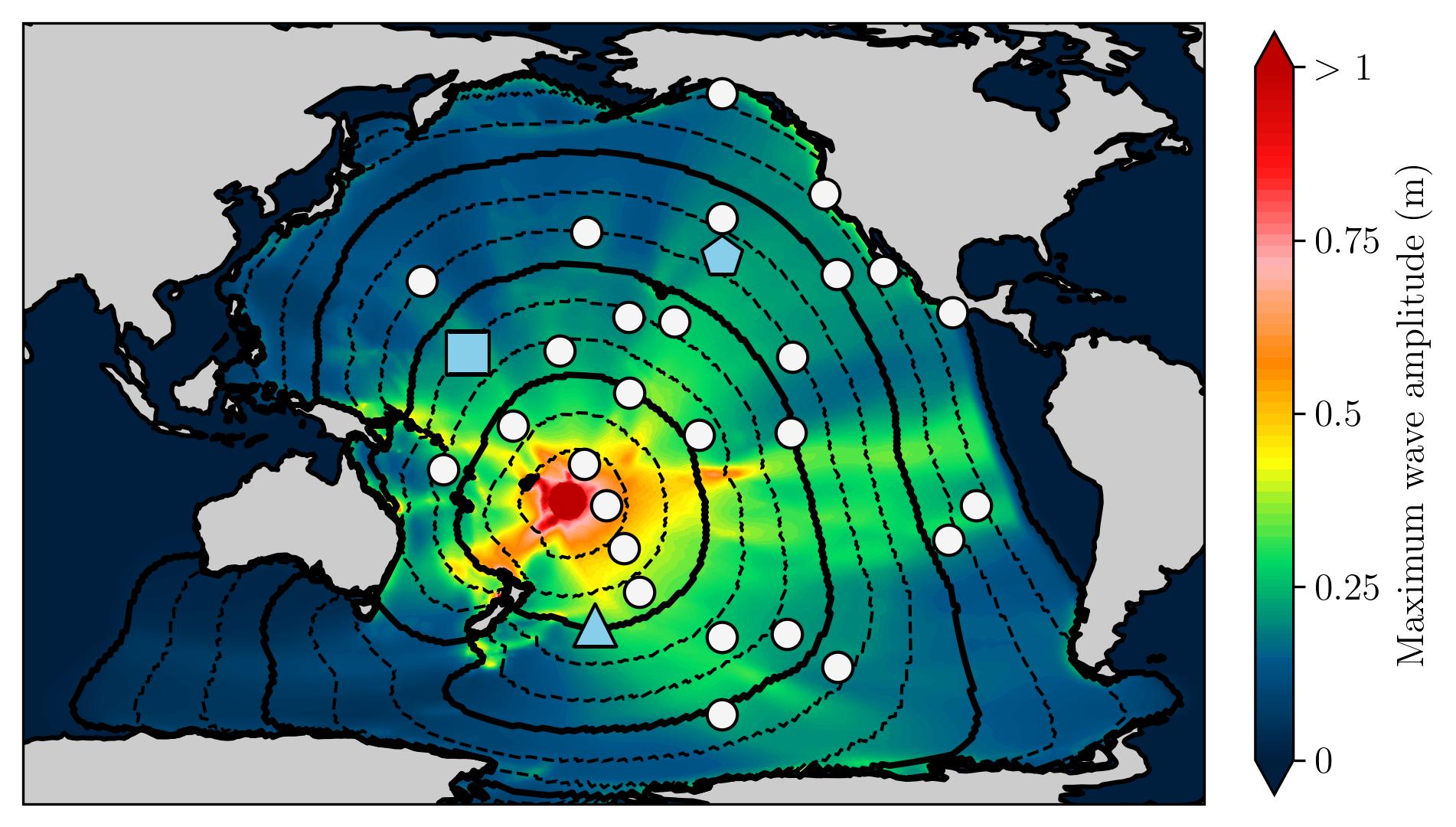}
\caption{\textit{Maximum amplitudes of the Fiji reference-case tsunami.} The maximum-amplitude field (top) shows the largest wave amplitude attained at each spatial location in the reference simulation. Curves indicate the wave-front position at hourly intervals (dashed) and every third hour (bold). Markers shown in blue indicate held-out fiducial sensor locations at which wave height is measured during the simulation, while white markers indicate sensor locations used to calibrate the randPROM.}
\label{fig:amplitude}
\end{figure}

The sensor domain \(\vtx\) is defined by 30 random grid points, sampled by rejection sampling using the time-integrated maximum amplitude as a target distribution and a uniform proposal. This favors sensor locations likely to experience significant wave activity. The maximum-amplitude field for the Fiji reference case is shown in Fig. \ref{fig:amplitude}. Also shown are 20 sampled sensor locations: 16 sensors (white markers) are used to calibrate the randPROM model, while 4 sensors are held out for testing (blue markers). This sensor domain determines the spatial locations at which the modes \(\Phi(\vtx)\) are evaluated.

\begin{figure}[h!]
\includegraphics[width=\textwidth]{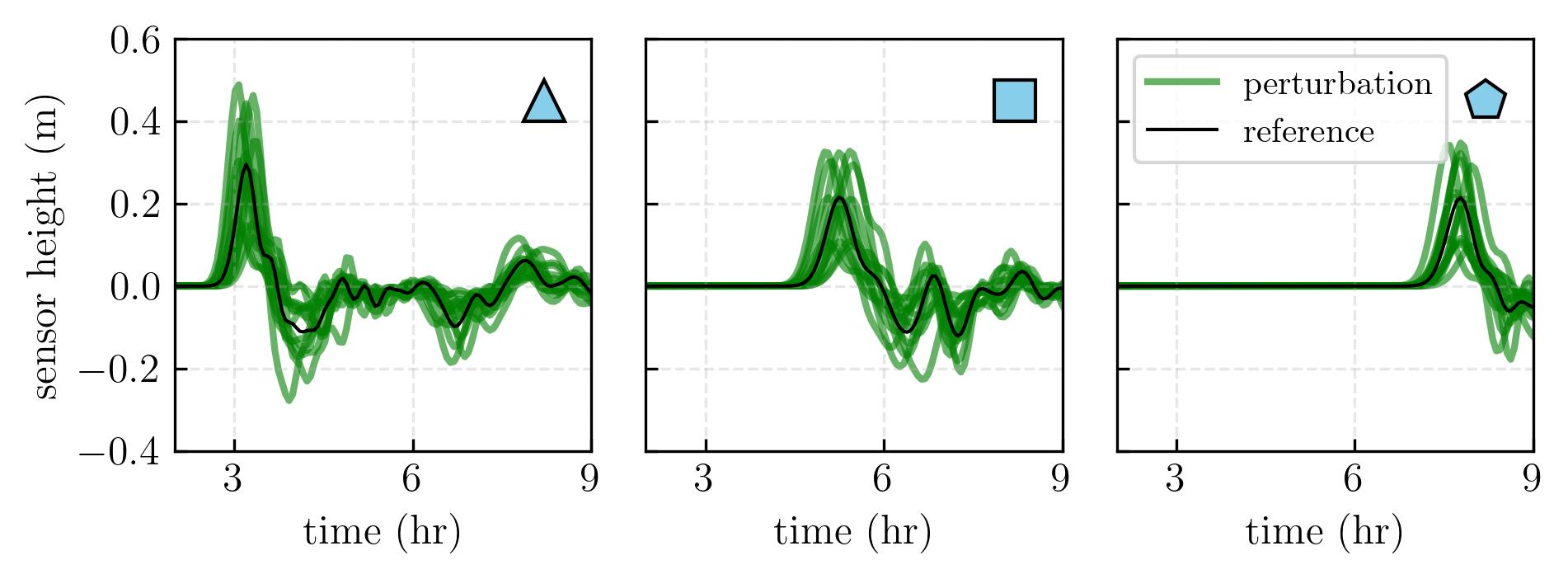}
\centering
\vspace{-0.5cm}
\caption{\textit{Sensor heights for the perturbation simulations.} Small changes in the initial condition relative to the reference simulation (black line) can produce substantially different wave heights at the sensors (green lines). These perturbation trajectories are used as calibration data for the randPROM model.}
\label{fig:sensor_perturbations}
\end{figure}

Although the perturbation simulations differ only slightly in initial condition, they can still yield substantial variation in both wave height and arrival time. This variability is illustrated for three example sensors in Fig. \ref{fig:sensor_perturbations}. Peak heights in the perturbation simulations (green curves) can approach nearly twice the peak height of the reference simulation (black curve), while arrival times may differ by more than 15 minutes before or after the reference.

\subsection{Demonstrating the Bayesian hierarchical calibration model}

\begin{figure}[h!]
\centering
\includegraphics[width=\textwidth]{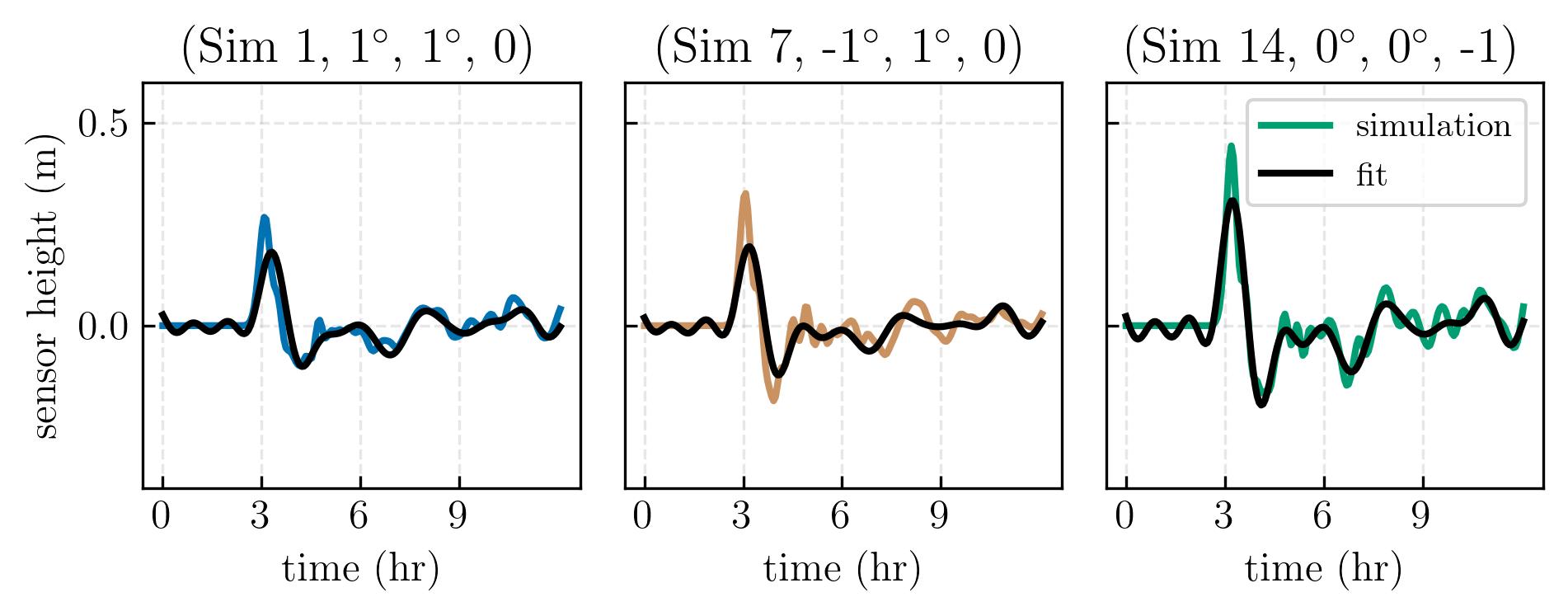}
\vspace{-0.5cm}
\caption{\textit{Example local posterior reconstructions for a randomly selected sensor near the epicenter.} Despite differing initial conditions, the randPROM produces highly accurate posterior fits (black curves) that agree closely with the corresponding simulation data at that sensor (colored curves). Because these are local posterior fits, they are conditioned on individual calibration scenarios; accordingly, we demonstrate them here on a representative high-activity sensor.}
\label{fig:randprom_fiji_local}
\end{figure}

The role of the hierarchical approach becomes clear when we examine how well posterior samples fit a given sensor trajectory for a calibration scenario \(\vb y_i\). By conditioning on a specific scenario while preserving the reduced dynamical structure, the randPROM produces physically consistent local posterior reconstructions (Fig. \ref{fig:randprom_fiji_local}). Despite the substantial variation across the calibration scenarios at a given sensor, the randPROM is able to fit each observed trajectory accurately.

\begin{figure}[h!]
\centering
\includegraphics[width=\textwidth]{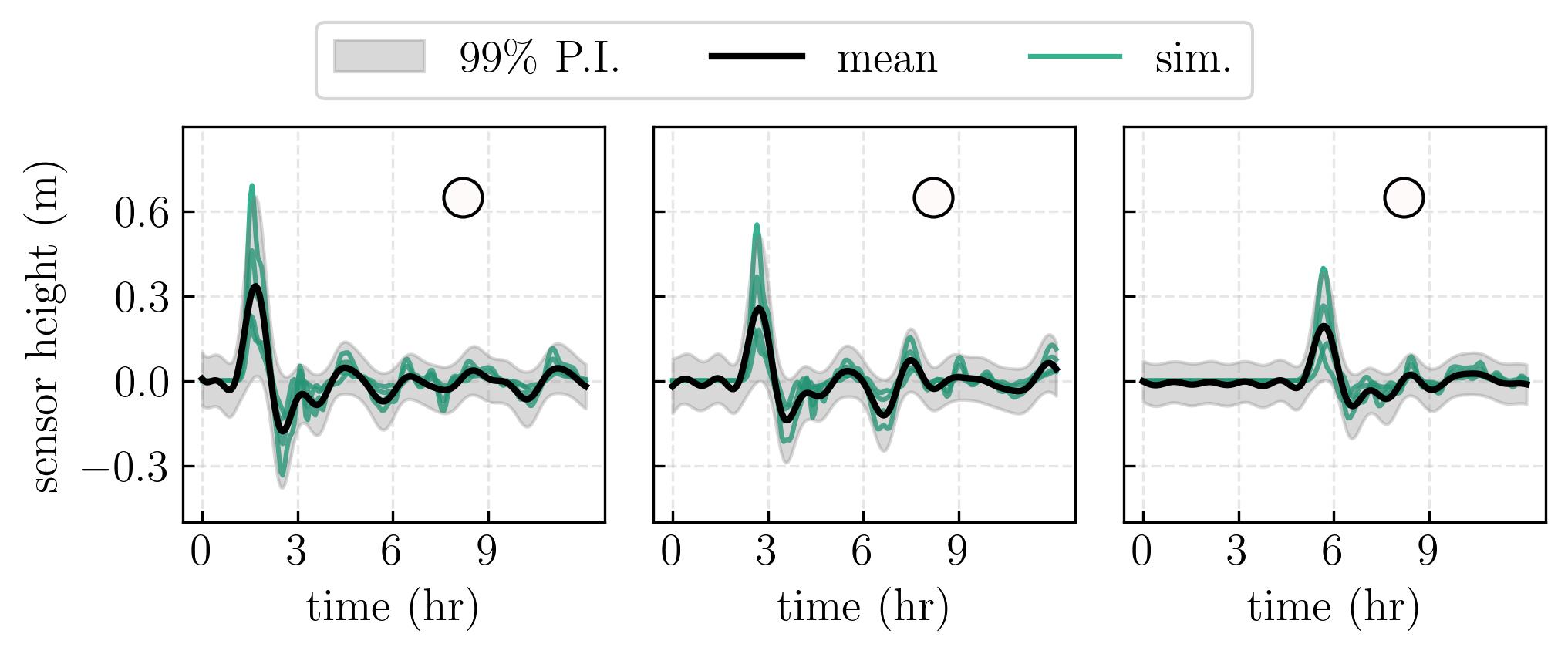}
\caption{\textit{Sensor reconstructions for three calibration scenarios in the training set using global posterior predictive summaries.} For three representative sensors, we show reconstructions based on the global randPROM together with 99\% credible intervals (CI) and prediction intervals (PI). These results illustrate the physical consistency and statistical interpretability of the global posterior predictive trajectories.}
\label{fig:randprom_fiji_global_in_train}
\end{figure}

Next, we illustrate the role of the global posterior predictive summaries in Fig. \ref{fig:randprom_fiji_global_in_train}. For each result, we draw samples from the final 1000 retained MCMC iterations and compute the 99\% posterior intervals with and without the observation-error term. These global summaries describe the range of behavior supported by the calibrated hierarchical model across the perturbation simulations, and the resulting credible intervals are broad enough to contain most of the corresponding local calibration trajectories.

\begin{figure}[h!]
\centering
\includegraphics[width=\textwidth]{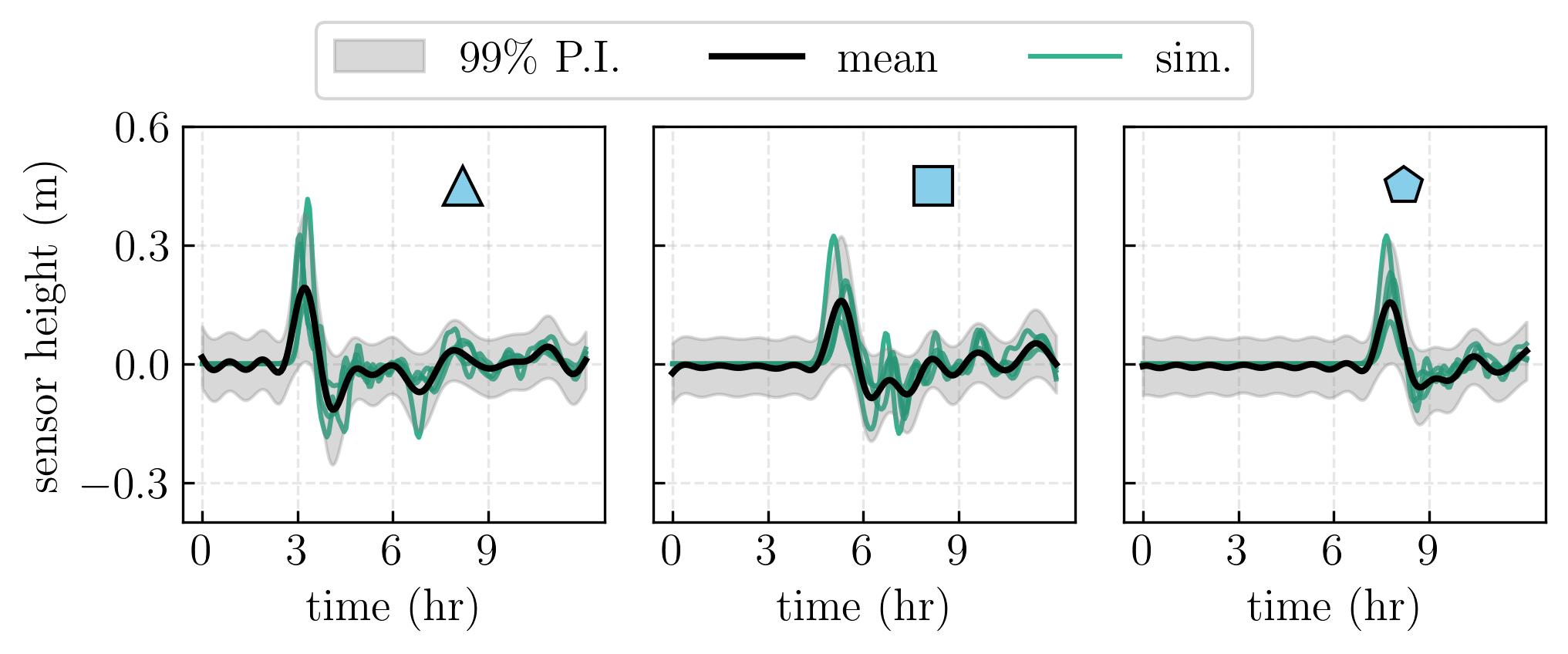}
\caption{\textit{Sensor reconstructions for three held-out sensors using global posterior predictive summaries.} As in Fig. \ref{fig:randprom_fiji_global_in_train}, we show sensor reconstructions based on the global hierarchical model, but now at sensors not used during calibration. This provides a first demonstration of the predictive setting explored further in subsequent sections.}
\label{fig:randprom_fiji_global}
\end{figure}

Furthermore, unlike the local posterior fits, the global posterior predictive summaries can be used to evaluate unseen sensors under the calibrated model. Figure \ref{fig:randprom_fiji_global} demonstrates this predictive capability of the randPROM. In a similar manner, the 99\% intervals capture both the overall trend and much of the variation observed in the perturbation data.

These results demonstrate the utility of the hierarchical calibration approach. Using local posterior distributions, we can accurately reconstruct sensor trajectories for each calibration scenario. Such local fits are particularly useful when conditioning on limited or early sensor data in real time, as considered later in Sec. \ref{sec:prediction_time} and Sec. \ref{sec:tohoku}. By contrast, the global posterior predictive summaries describe the broader family of behaviors supported by the calibrated model, but remain limited by the variability represented within the calibration scenarios. Extending this framework to support broader predictive generalization requires additional structure, which we study in the following section.

\subsection{Prediction using fewer sensors and fewer time points}
\label{sec:prediction_time}

A key advantage of the randPROM framework is that, even when calibrated with sparse data, it can be used to predict the remaining temporal domain and to infer wave behavior at unobserved sensors. The framework also naturally accommodates missingness, whether from isolated missing observations in time or the complete loss of a sensor record. In the Fiji study, the full simulation spans 12 hours, but in the present analysis we restrict attention to the first 9 hours. Motivated by this, we perform an additional study in which the MCMC calibration uses only 20\% and 50\% of the temporal domain from the calibration scenarios. In these reduced-window runs, calibration sensors are restricted to those whose absolute wave-height signal in the reference simulation exceeds 0.05 within the retained time window, while the plotted sensors are selected from the held-out set. The goal is to assess the quality of the resulting predictions under reduced temporal coverage.

\begin{figure}[h!]
\includegraphics[width=\textwidth]{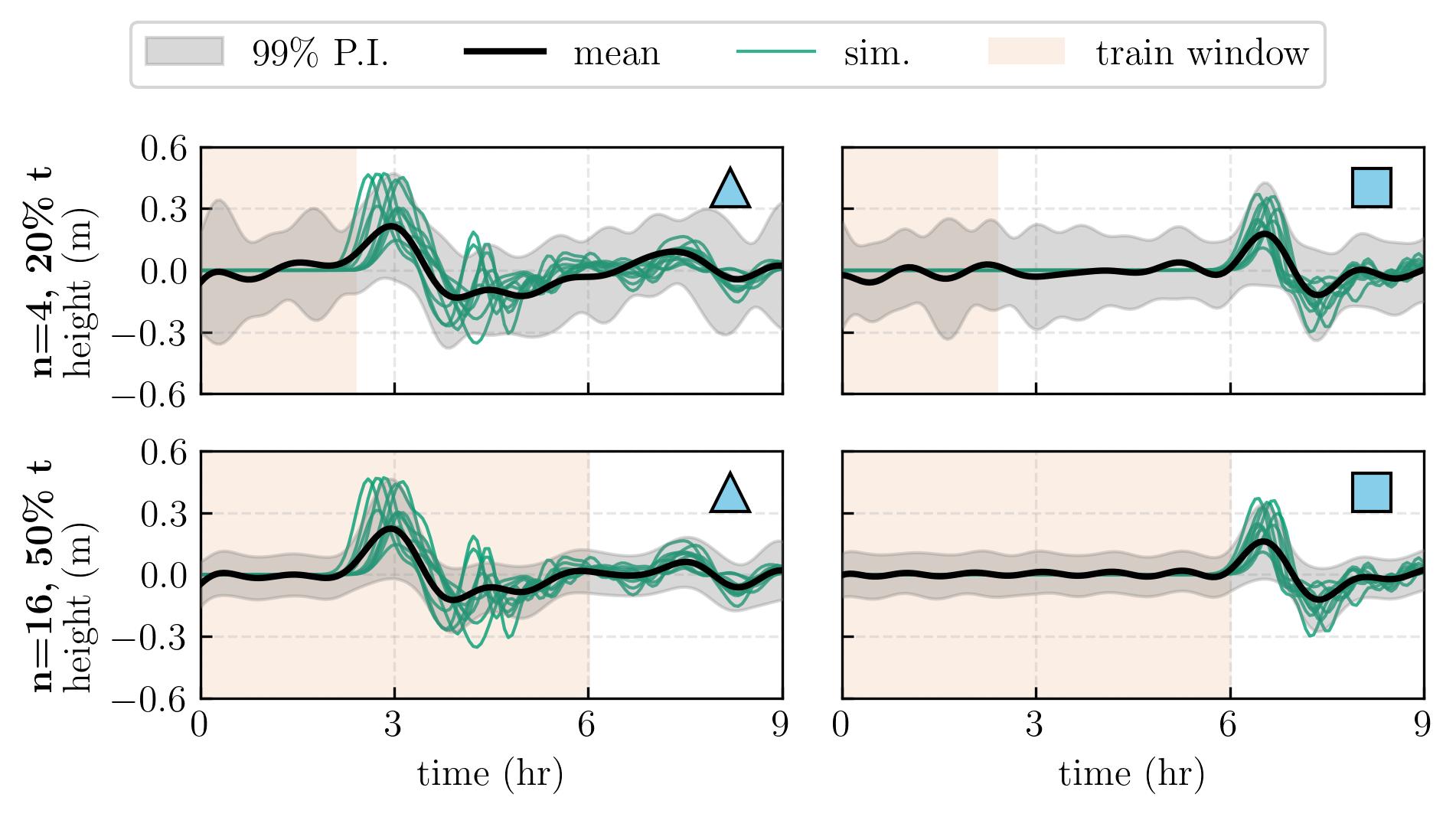}
\centering
\vspace{-0.5cm}
\caption{\textit{Sensor predictions using a truncated temporal domain.} We demonstrate the predictive capabilities of randPROM when the available temporal data are limited to 20\% and 50\% of the 9-hour analysis window. Each row shows predictions under this restriction, with the 99\% prediction intervals (gray regions) compared against simulated sensor trajectories (red curves).}
\label{fig:fiji_global_low_time}
\end{figure}

The results shown in Fig. \ref{fig:fiji_global_low_time} indicate that randPROM remains stable even under substantially reduced temporal coverage. For visualization, the predictive summaries shown in Fig. 13 are reconstructed on the full sensor basis from the saved posterior states, so that the held-out sensors can be evaluated directly. Using shortened temporal windows together with the sensors that satisfy this amplitude-based activity criterion, the model continues to produce predictions that remain consistent with the simulated sensor trajectories. Remarkably, the 20\% and 50\% cases remain very similar even after enforcing this active-sensor restriction, suggesting that the reduced model and hierarchical prior already provide substantial regularization once the most informative sensors are included.

\section{RandPROM modeling of the 2011 Tohoku tsunami}
\label{sec:tohoku}

An important practical goal of the randPROM methodology is its application to real-world tsunami events. The preceding Fiji experiments suggest that such an approach may be viable. In operational settings, tsunami inference relies on sparse wave sensors that become active only as the wave propagates across the ocean. The goal is to infer event characteristics as early as possible, so that forecasts can be issued quickly and accurately. This motivates the analysis in Sec. \ref{sec:prediction_time}, where randPROMs were calibrated using only early-time data from a limited number of active sensors. More broadly, estimating the initial conditions of tsunami-generating earthquakes is a detailed and time-consuming inverse problem that may require months or even years of analysis. In practice, one must therefore rely on lower-order initial estimates for near-real-time forecasting, while using available data to calibrate uncertainty in the resulting predictions.

\begin{figure}[t!]
\centering
\includegraphics[width=\textwidth]{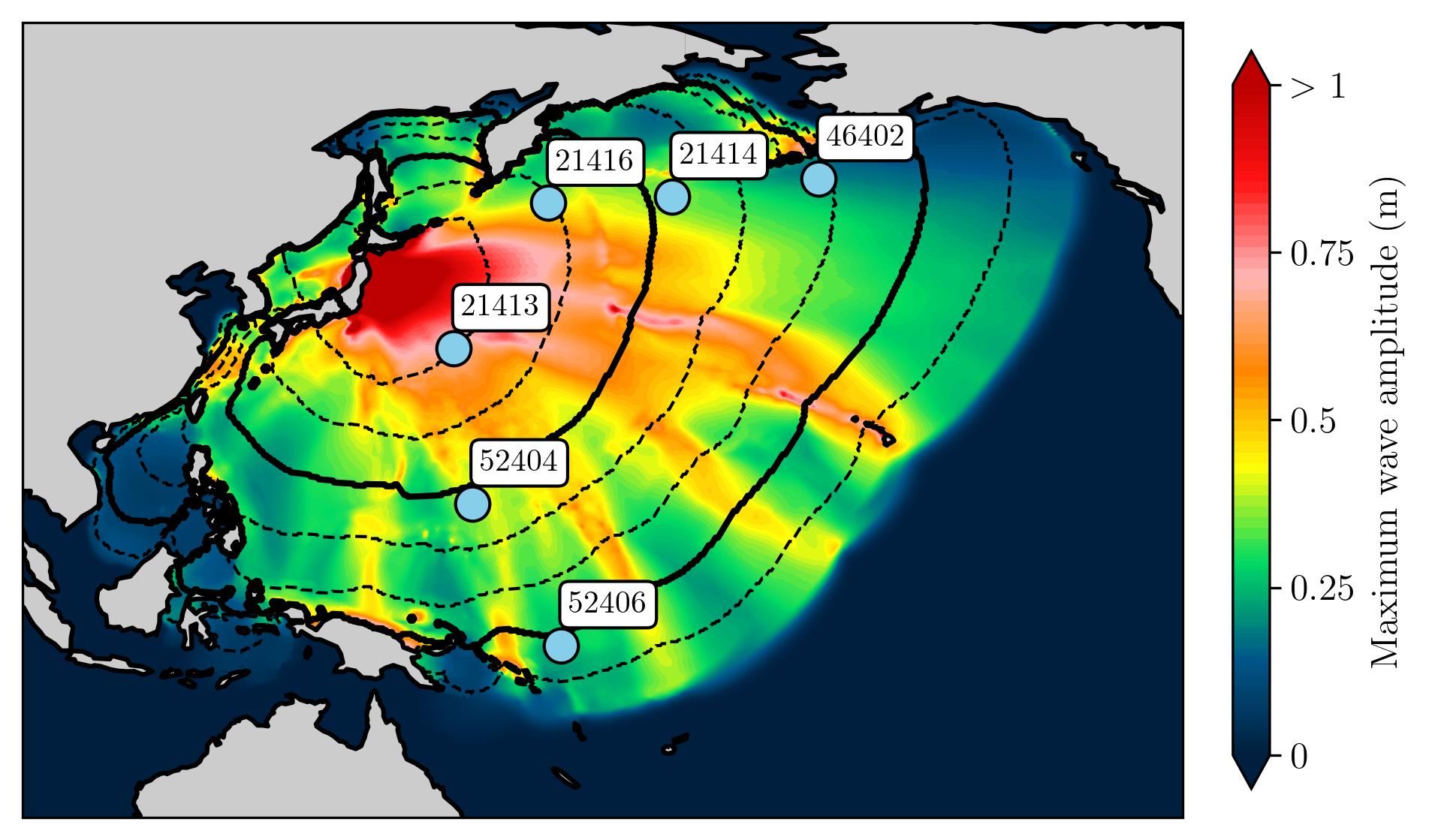}
\caption{\textit{Amplitude field and sensor locations for the Tohoku tsunami.} The plotting style follows the Fiji amplitude-field visualization in Fig. \ref{fig:amplitude}. The sensor locations correspond to the DART buoys used for validation.}
\label{fig:tohoku_amp}
\end{figure}

In this section, we apply the methodology of the previous sections to the 2011 Tohoku tsunami. The Tohoku event \cite{fujii2011tsunami} resulted from a magnitude 9.0 earthquake near the Japan Trench at $38.1035^\circ\mathrm{N}$, $142.861^\circ\mathrm{E}$ and caused extensive flooding and damage in Japan, including at the Fukushima Daiichi nuclear power plant. The shallow-water-equation solver used in this study has been validated previously against the Tohoku tsunami. For the highest-fidelity simulation, the initial conditions are given by an elevated sea-surface displacement inferred through waveform inversion \cite{fujii2011tsunami}. Additional details of the simulation setup are provided in \cite{mcdugald2025attention}.

\begin{figure}[ht!]
\centering
\includegraphics[width=\textwidth]{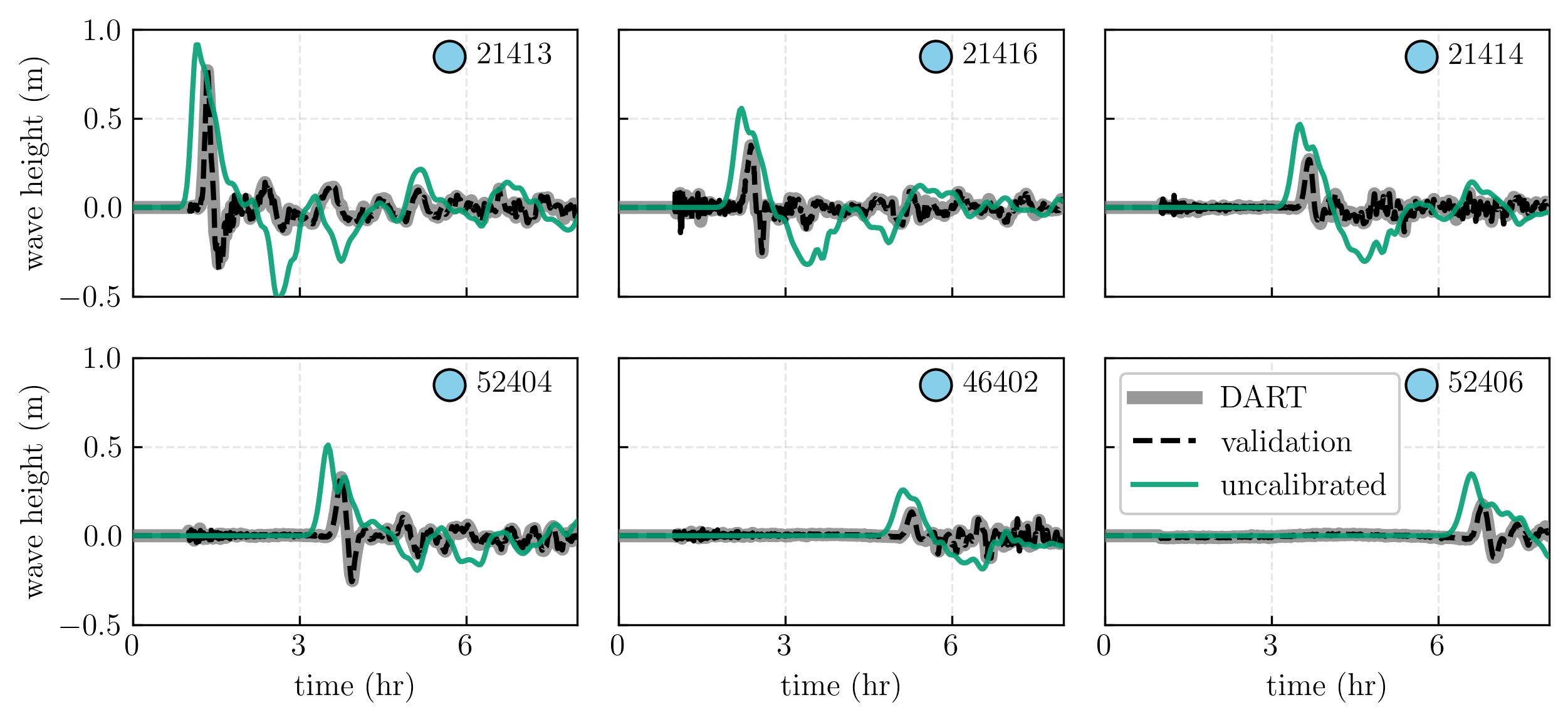}
\caption{\textit{Sensor readings for the Tohoku tsunami.} DART buoy observations (black) show the wave heights recorded during the Tohoku event. The highest-fidelity validation simulation, based on carefully inferred initial conditions (black dashed), shows strong agreement with the historical data. A lower-fidelity uncalibrated simulation using a coarse Gaussian initial condition (green) captures the broad arrival-time behavior but agrees less well overall.}
\label{fig:tohoku_sensors}
\end{figure}

The resulting simulated wave heights are compared directly against DART buoy observations of the event. These validation results are summarized in Fig. \ref{fig:tohoku_amp}, beginning with the amplitude field of the validated high-fidelity simulation and the locations of the DART sensors used for comparison. The corresponding sensor time series are shown in Fig. \ref{fig:tohoku_sensors}, where we compare the observed event, the validated simulation, and an uncalibrated lower-fidelity simulation based on a naive Gaussian surface deflection.

These two sets of initial conditions differ substantially, and the Gaussian initial condition reproduces only the broad qualitative behavior of the event, such as approximate arrival times, while failing to match the observed amplitudes accurately. Nevertheless, the corresponding randPROM can be calibrated to better represent the resulting model discrepancy. In realistic forecasting settings, the precise initial conditions are not available quickly, both because of the detailed inverse analysis required and because of the computational expense of validating high-resolution simulations. For example, the detailed initial conditions used here for validation were not reported until approximately six months after the event \cite{fujii2011tsunami}. Near real-time forecasting must therefore rely on simpler initial estimates. The relevant question is then how to quantify uncertainty and produce informative error bounds when starting from such lower-order approximations. The present method aims to show that randPROMs can be calibrated to provide such probabilistic corrections.

\begin{figure}[ht!]
\centering
\includegraphics[width=\textwidth]{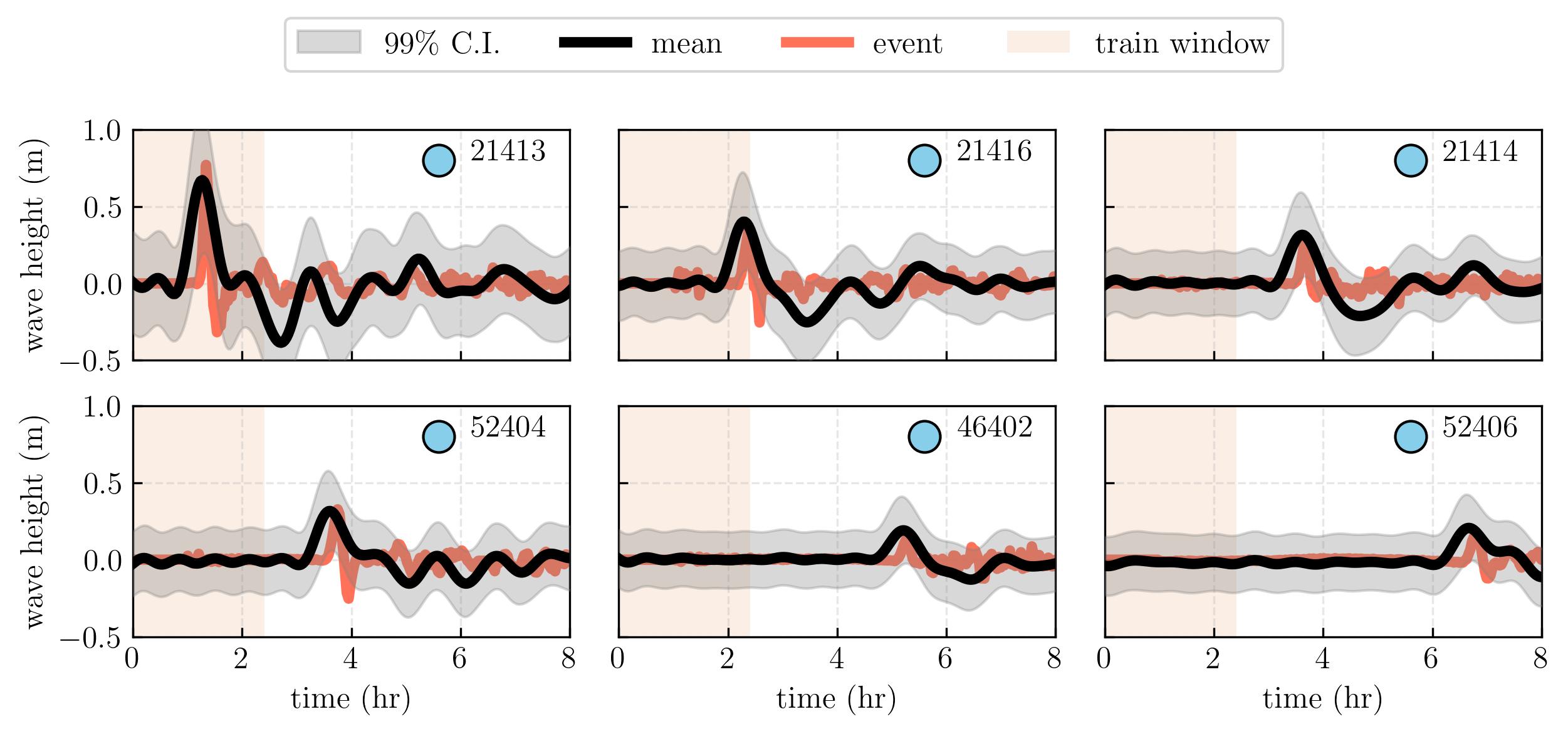}
\caption{\textit{Sensor reconstructions for the Tohoku case.} Calibrating the lower-fidelity model to the historical data improves the reconstruction of the sensor trajectories shown in Fig. \ref{fig:tohoku_sensors}.}
\label{fig:tohoku_sensor_recon}
\end{figure}

\begin{figure}[ht!]
\centering
\includegraphics[width=\textwidth]{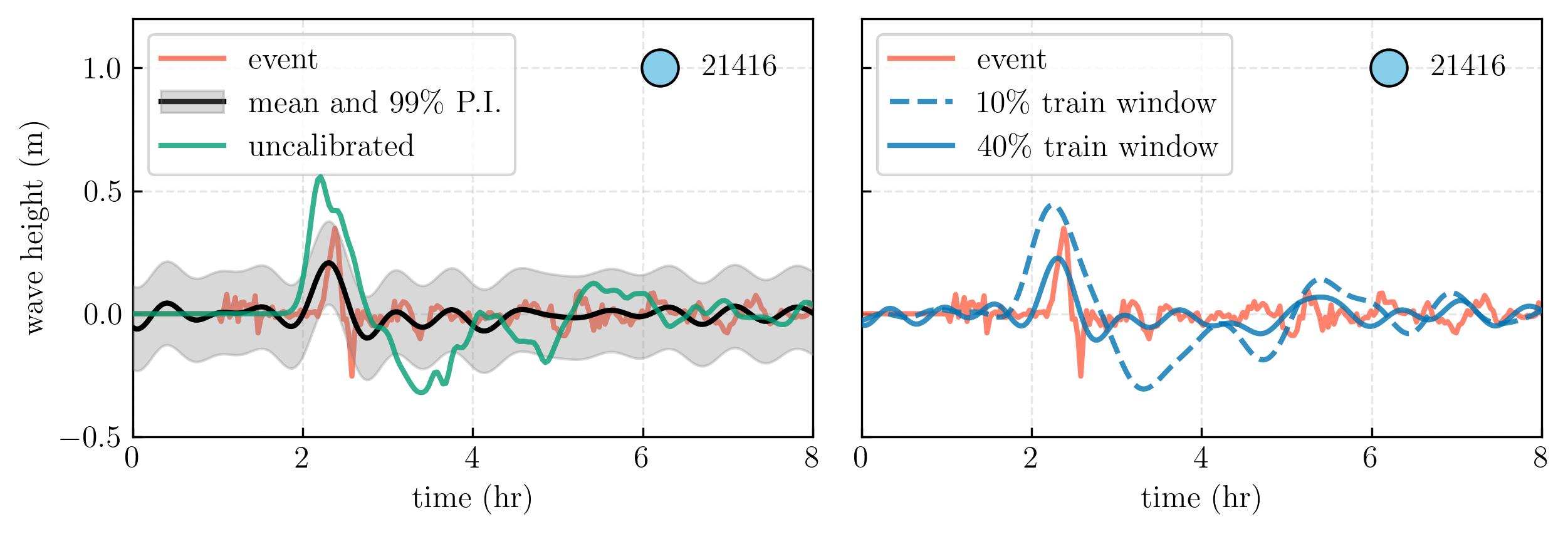}
\caption{\textit{Detailed reconstruction for the Tohoku case at sensor \#21416.} The calibrated model shows improved accuracy relative to the uncalibrated initial guess. This close-up illustrates how the posterior predictive reconstruction using 100\% of the temporal domain suppresses inaccurate wave-height trajectories, with performance improving as additional sensor data are incorporated (right).}
\label{fig:tohoku_single_pct}
\end{figure}

We begin by constructing a corrected reduced-order model from the simulation initialized with the Gaussian surface displacement. For the calibration scenarios required by the corresponding randPROM, all six DART sensors shown in Fig. \ref{fig:tohoku_amp} are used. We then apply the Bayesian hierarchical calibration model described in Sec. \ref{sec:bayesian}, with the expectation that, \textit{a priori}, the simplified Gaussian initial condition and its associated reduced initial values will only approximate the observed Tohoku event coarsely. Nevertheless, there may exist event-specific latent initial values that lead to substantially improved agreement with the observed data, while the hierarchical posterior distribution quantifies the corresponding uncertainty. The results of this calibration, summarized through posterior predictive reconstructions, are shown in Fig. \ref{fig:tohoku_single_pct}.

These results build on the findings of Sec. \ref{sec:prediction_time}, namely that relatively limited early-time data can still support useful reconstruction and prediction. In the Tohoku case, we analyze the first 8 hours of the event and consider truncated temporal windows so that calibration relies primarily on early-arrival information. Even under this restriction, the posterior predictive mean captures much of the observed Tohoku behavior across the sensor network. Late-time reconstruction error at sensor \#21413 remains associated with substantial uncertainty, but the posterior mean still represents the low-wave-activity regime well. The remaining sensor predictions remain qualitatively consistent with the observed trends.

Furthermore, relative to the uncalibrated lower-fidelity model, the calibrated randPROM reduces systematic over- and under-prediction in several of the displayed sensor traces, owing to the regularization introduced by the hierarchical pooling structure. In particular, the posterior predictive trajectories better capture the wave peaks observed at several sensors. This effect is shown for sensor \#21416 in Fig. \ref{fig:tohoku_single_pct}: in the left panel, the event-conditioned posterior reconstruction provides a closer fit to the observed event (red line) than the uncalibrated model, while in the right panel we show that the posterior mean remains informative even when calibrating with only 10\% of the temporal domain or, alternatively, with 40\% of the temporal domain.

These findings suggest that calibration of a randPROM can materially alter a crude initial simulation, allowing the reduced model to capture more of the observed event structure than is present in the original lower-fidelity forecast. They also provide a second demonstration that the approach remains informative when only sparse early-time data are available.

\subsection{Quantitative Summary}
\label{sec:quantitative_summary}

To complement the discussions and plots of Figs. 11--17, Table \ref{tab:metrics} summarizes quantitative error and uncertainty metrics for the representative Fiji and Tohoku results. The Fiji rows corresponding to Fig. 13 are reported for the 20\% and 50\% temporal-window cases, respectively, while the Tohoku rows corresponding to Fig. 17 are reported for the 100\%, 10\%, and 40\% temporal-window cases, respectively.

\begin{table}[t!]
\centering
\caption{\textit{Quantitative summary metrics for representative Fiji and Tohoku results.}
For each case, we report the mean relative \(L^2\) error, the mean peak-height error, and the empirical coverage of the nominal 99\% prediction interval (PI).}
\label{tab:metrics}
\begin{tabular}{lllrrr}
\toprule
Case & Figure & \% $t$ & Mean rel.\ $L^2$ & Mean peak err.\ (m) & 99\% PI cov. \\
\midrule
Fiji & Fig. 11 & 100 & 0.6552 & 0.1552 & 0.9910 \\
Fiji & Fig. 12 & 100 & 0.6932 & 0.1021 & 0.9886 \\
Fiji & Fig. 13 & 20 & 0.8448 & 0.1128 & 0.9980 \\
Fiji & Fig. 13 & 50 & 0.6905 & 0.1141 & 0.9902 \\
Tohoku & Fig. 16 & 30 & 1.8962 & 0.0508 & 0.9882 \\
Tohoku & Fig. 17 & 100 & 0.9221 & 0.1395 & 0.9959 \\
Tohoku & Fig. 17 & 10 & 2.5781 & 0.0964 & 0.7178 \\
Tohoku & Fig. 17 & 40 & 1.0900 & 0.1205 & 0.9917 \\
\bottomrule
\end{tabular}
\end{table}

These metrics help distinguish two complementary aspects of predictive quality. The mean relative \(L^2\) error measures agreement over the entire sensor trajectory, whereas the mean peak-height error isolates the fidelity of the largest observed wave height. The 99\% PI coverage quantifies how the posterior uncertainty bands are calibrated with respect to the observed sensor measurements.

For the Fiji examples, the results are consistent with the visual reconstructions: interval coverage remains close to nominal, and the trajectory and peak errors remain moderate across both the full-window and reduced-window studies. In particular, the two Fig. 13 rows remain close even under the corrected active-sensor restriction, reinforcing the observation that the randPROM remains stable under substantially reduced temporal coverage once more calibration sensors are included.

For the Tohoku examples, the peak-height errors remain small, and the posterior intervals are generally well-calibrated once a sufficient fraction of the early-time record is included. The larger relative \(L^2\) errors should be interpreted with some care, since the observed DART traces contain long low-amplitude intervals for which modest absolute discrepancies can still produce large relative trajectory error. In addition, the earliest truncated windows may favor a larger initial-amplitude reconstruction because only the first wave activity is being represented. As more of the temporal record is incorporated, the posterior prediction must accommodate additional downstream activity, which damps the initial peak and can slightly increase the mean peak-height error even as the 99\% PI coverage improves substantially (visually covering both the peak and low-wave activity in the sensor observations). In this setting, the peak-height error and interval coverage provide an important complement to the relative \(L^2\) metric when assessing predictive quality.

\section{Discussion}

A key capability of the randPROM framework, and an important advantage over POD alone, is that the corrected reduced-order model may be evolved from \textit{any} initial reduced state ($\vb a_0^\star$), thereby generating new reduced trajectories, e.g.,
\begin{equation*}
g(\vb a_0^\star,t)=\vb a^\star(t).
\end{equation*}
If these trajectories remain valid representations of plausible variations of the reference simulation, then the corrected reduced model can serve as a surrogate for generating new scenarios. However, no \textit{explicit} mapping currently exists between the physical simulation inputs (for example, earthquake epicenter and magnitude) and the corresponding reduced initial conditions. Developing such a mapping would be an important next step, as it could reduce reliance on early sensor calibration and move the predictive window closer to the time of the seismic event. The central modeling challenge is therefore to learn a statistical distribution over reduced initial conditions that can be calibrated to new inputs ($\tilde{\vb x}$), and to verify that the resulting reduced trajectories remain physically meaningful. When successful, this yields the probabilistic surrogate model we refer to as a randPROM. In this way, the task of probabilistically modeling the full reduced trajectory is reduced to inferring only the initial reduced coefficients $\vb a_0$.

\subsection{Real-time utility and social responsibility statement}

As noted earlier, the ultimate goal of a tsunami forecast model is to provide key hazard characteristics, including wave arrival time, wave height, and inundation extent, for all affected regions. The present randPROM analysis investigates only the first two of these quantities.

In practical settings, a tsunami simulation initialized with a simple Gaussian disturbance may take on the order of 2 minutes to run on a modern laptop, whereas a higher-fidelity simulation may require roughly 20 minutes. Yet, in order to characterize uncertainty, one may need tens, hundreds, or even thousands of such simulations. A randPROM constructed from a reference solution may alleviate some of this cost by replacing repeated forward simulations with a calibrated reduced surrogate over nearby scenarios. Its potential utility as a reduced-order model is therefore substantial, both in terms of spatiotemporal compression and computational efficiency.

At present, the technology is not yet intended to fully replace conventional workflows end-to-end, although the current implementation already operates on a comparable practical timescale. In practice, the full procedure described here can take on the order of 10 minutes, as observed for both the Fiji and Tohoku cases. As training procedures become faster and more reliable, we expect this framework to become increasingly useful as a parallel approximation tool. Importantly, much of the work required to build reference reduced models over neighborhoods of plausible events can be performed offline and cached in advance, so that calibrated local randPROMs may be deployed rapidly once an earthquake occurs. 

A remaining practical challenge is that the usefulness of cached reference neighborhoods depends on how well they cover the space of plausible events.  The reduced models themselves are relatively compact; for the Fiji case the reduction is on the order of 250× to 300× smaller in storage than the corresponding neighborhood of full simulation histories. Thus the primary scalability issue is not storage but the construction of a sufficiently representative event library, together with improved mechanisms for transferring information across related events, such as improved basis selection and transfer-learning strategies.

\subsection{Future work}

This work makes two main contributions. First, it investigates reduced-order modeling for a challenging nonlinear system, namely the shallow-water equations, in a setting that requires relatively many retained modes, model simplifications, and additional care in constructing corrected reduced models. This includes the operator-level calibration of both the linear and quadratic reduced operators. Second, the randPROM framework combines a projection-based reduced-order surrogate with a Bayesian hierarchical calibration model. As a result, there are many opportunities to improve both components of the methodology and move the framework toward operational relevance.

On the reduced-order modeling side, there may be substantial room for improvement in the generation of modes. Although POD remains a standard choice in projection-based ROMs, alternative decompositions such as dynamic mode decomposition may also be useful \cite{kutz2016dynamic}. In addition, emerging approaches such as $\beta$-variational autoencoders may provide learned latent structures with more desirable properties \cite{solera2024beta}. We intend to investigate whether alternative reduced representations can produce modes and coefficients that transfer more effectively across related scenarios.

Several improvements are also possible in the training of the corrected reduced model. These include better optimization strategies, such as learning-rate scheduling and transfer learning, as well as implementation of ODE solvers specialized for the reduced dynamics. Stable, accurate, and efficient training remains critical if the method is to be deployed in practical settings.

Within the Bayesian hierarchical framework, several extensions could improve both inference and prediction. For example, the observation-noise model could be made heteroskedastic, allowing uncertainty to vary with time and signal amplitude rather than remaining constant across the full trajectory. Such an extension of Eq. \ref{eq:fullfull} would permit larger uncertainty near wave peaks and smaller uncertainty during quiescent periods. Similarly, asymmetric or weighted likelihood formulations may improve calibration in the portions of the signal that are most operationally important.

Finally, validation against real-world data remains essential. Additional studies using historical tsunami sensor data are needed to assess the practical reliability of the method. This also includes improved treatment of initial conditions for complex real-world events, as suggested by the present Tohoku case study.

\section*{Open Research Section}
All data, code, and supporting documentation required to reproduce the analyses are publicly available \cite{coffing2026reduced}. The tsunami simulation data used in this work can be generated from the open source software \textit{swe-python} \cite{swegithub}.

\section*{Conflict of Interest declaration}
The authors declare there are no conflicts of interest for this manuscript.

\acknowledgments
We thank the reviewers for the detailed reading and insightful comments, which greatly improved the quality and clarity of this work. This work was supported by the U.S. Department of Energy through the Los Alamos National Laboratory. Los Alamos National Laboratory is operated by Triad National Security, LLC, for the National Nuclear Security Administration of U.S. Department of Energy (Contract No. 89233218CNA000001).

\bibliography{refs}

\end{document}